\documentclass[sn-mathphys-num]{sn-jnl}

\usepackage{graphicx}%
\usepackage{multirow}%
\usepackage{amsmath,amssymb,amsfonts}%
\usepackage{amsthm}%
\usepackage{mathrsfs}%
\usepackage[title]{appendix}%
\usepackage{xcolor}%
\usepackage{textcomp}%
\usepackage{manyfoot}%
\usepackage{booktabs}%
\usepackage{adjustbox}
\usepackage{algorithm}%
\usepackage{algorithmicx}%
\usepackage{algpseudocode}%
\usepackage{listings}%
\usepackage{caption}
\usepackage{subcaption} 
\usepackage{graphicx} 
\usepackage{array}
\usepackage{anyfontsize}
\usepackage{float}
\usepackage{tabularx}
\usepackage{array} 

\usepackage[most]{tcolorbox}
\usepackage[colorinlistoftodos]{todonotes}
\usepackage{adjustbox}
\usetikzlibrary{shadows}

\definecolor{milkgreen}{rgb}{0.5,1,0.5}

\newtcolorbox{myshadowbox}{
  colback=blue!30, 
  colframe=blue!40, 
  boxrule=1pt, 
  boxsep=0pt, 
  left=8pt, right=8pt, top=6pt, bottom=6pt,
  enhanced, 
  drop shadow={gray!50, opacity=0.5, shadow xshift=0.8em, shadow yshift=-0.8em}
}

\renewcommand\tabularxcolumn[1]{m{#1}} 

\theoremstyle{plain}

\theoremstyle{definition} 
\begin{document}

\title[Article Title]{FairAIED: Navigating Fairness, Bias, and Ethics in Educational AI Applications}

\author*[1]{\fnm{Zhipeng}\sur{Yin}}\email{zyin007@fiu.edu}
\author[1]{\fnm{Sribala Vidyadhari}\sur{Chinta}}\email{schin079@fiu.edu}
\author[1]{\fnm{Zichong}\sur{Wang}}\email{zwang114@fiu.edu}
\author[2]{\fnm{Matthew}\sur{Gonzalez}}\email{0588830@students.dadeschools.net}
\author*[1]{\fnm{Wenbin}\sur{Zhang}}\email{wenbin.zhang@fiu.edu}

\affil[1]{\orgname{Florida International University}}
\affil[2]{\orgname{TERRA Environmental Research Institute}}
\abstract{
    The integration of Artificial Intelligence (AI) in education holds immense potential for personalizing learning experiences and transforming instructional practices. However, AI systems can inadvertently encode and amplify biases present in educational data, leading to unfair or discriminatory outcomes. As researchers have sought to understand and mitigate these biases, a growing body of work has emerged examining fairness in educational AI. These studies, though expanding rapidly, remains fragmented due to differing assumptions, methodologies, and application contexts. Moreover, existing surveys either focus on algorithmic fairness without educational setting or emphasize educational methods while overlooking fairness. To address these concerns, this survey provides a comprehensive and systematic review of algorithmic fairness within educational AI, explicitly bridging the gap between technical fairness research and educational applications. We integrate multiple dimensions, including bias sources, fairness definitions, mitigation strategies, evaluation resources, and ethical considerations, into a harmonized, education-centered framework. Furthermore, we extend existing fairness taxonomies by incorporating a newly introduced multi‑level dimension alongside the traditional individual‑ and group‑level dimensions, capturing fairness concerns that simultaneously affect both individual students and student groups. In addition, we explicitly examine practical challenges such as censored or partially observed learning outcomes and the persistent difficulty in quantifying and managing the trade-off between fairness and predictive utility, enhancing the applicability of fairness frameworks to real-world educational AI systems. Finally, we outline an emerging pathway toward equitable AI-driven education and by situating these technologies and practical insights within broader educational and ethical contexts, this review establishes a comprehensive foundation for advancing fairness, accountability, and inclusivity in the field of AI education.
}


\maketitle

\section{Introduction}

Artificial intelligence (AI) has emerged as a transformative force, significantly reshaping the methodologies for learning and teaching. The historical trajectory of AI in education (AIED) can be traced back to pioneering experiments with computer-assisted instruction in the 1960s and 1970s, closely paralleling the broader evolution of AI that began with the seminal Dartmouth Workshop in 1956~\cite{springer2022,fullestop2024personalized,hbr2020ai}. Early implementations sought to tailor technology-based instruction to individual learning styles and paces, a foundational concept substantially advanced through contemporary AI techniques~\cite{princetonreview2024ai}. Today, AI's transformative role is exemplified through personalized learning experiences, automated grading systems, and language translation services, collectively addressing diverse educational needs~\cite{datasciencecentral2024automated,princetonreview2024ai,topmarks2024how}. For example, AI-powered educational platforms dynamically adapt instructional content, enhancing student engagement by generating individualized learning paths in real-time based on students' goals, interests, and prior knowledge~\cite{mciu2024addressing}. Furthermore, automation in assessment processes, demonstrated by systems such as Graide and Top Marks AI, provides consistency and detailed feedback, substantially reducing educator workloads and allowing greater focus on complex instructional tasks~\cite{baker2022intertwined,elearningindustry2024ai}. Similarly, AI-driven language translation and support services have also contributed significantly to breaking down language barriers in education, enabling learners worldwide to access educational resources.

Despite these significant advances in educational AI, its integration into learning environments continues to raise critical concerns about fairness, bias, and ethical accountability~\cite{ferrara2023fairness,pressbooks2022}. In particular, AI systems can unintentionally perpetuate biases inherent in their training data, disproportionately impacting historically marginalized student groups~\cite{MonashEdu}. For instance, essay grading system may embed biases reflective of subjective preferences of human graders, disadvantaging students whose writing styles or cultural backgrounds deviate from established norms~\cite{princetonreview2024ai,luckin2022brief}. Similarly, AI-based early-warning systems intended to identify at-risk students often overpredict risk for certain demographic groups, such as students of lower socioeconomic backgrounds, because of historical bias embedded in attendance or disciplinary data~\cite{Zhao2017MenAlsoLikeShopping}. In addition, adaptive learning platforms designed to personalize instruction, while beneficial, may inadvertently reinforce learning disparities by providing fewer challenging materials to underperforming students, thereby limiting their academic growth and perpetuating stereotype bias~\cite{Siemens2015LearningAnalytics,Baker2022AlgorithmicBias}.

Recognizing these critical challenges, existing research has explored diverse strategies to enhance fairness and mitigate bias in educational AI systems. For instance, Hu and Rangwala~\cite{hu2020towards} examined methods that ensure equitable treatment for students with comparable learning profiles, whereas Liu \textit{et al.}~\cite{liu2025leveraging} proposed approaches that promote comparable opportunities across student populations with differing backgrounds. Despite these advancements, prior studies rely on distinct assumptions, fairness dimensions, and application contexts, making it difficult to establish a harmonized understanding of fairness across educational settings. Although several surveys~\cite{leslie2024ai,bellamy2019ai,pessach2023algorithmic,varona2022discrimination} exist within the broader AI fairness literature, they primarily emphasize algorithmic or technical aspects without grounding fairness in the educational context. Conversely, educational AI surveys~\cite{xu2022systematic,luckin2019designing,ayeni2024ai,pedro2019artificial} tend to focus on pedagogical methodologies, system design, or learning outcomes, often overlooking fairness as a central analytical lens. Furthermore, most fairness-related surveys~\cite{mehrabi2021survey,ferrara2024fairness,richardson2021framework} neglect the multi-level nature of fairness, where biases emerge simultaneously at individual and group levels, as well as key practical challenges, including censored or partially observed learning outcomes and the absence of unified protocols for determining appropriate fairness-utility selection in educational AI application. These issues render many existing fairness frameworks, which assume full label availability and clear utility measures, inapplicable or inadequate for real-world educational scenarios. The lack of unified evaluation protocols makes it difficult for practitioners to quantify and control trade-offs between fairness and predictive utility. Consequently, there remains a pressing need for a comprehensive and practical survey that integrates these dimensions into a coherent education-centered framework, guiding researchers and practitioners toward equitable and ethically grounded AI-driven educational systems.

To this end, this survey provides a comprehensive and systematic review of fairness in educational AI, explicitly addressing both the fragmented nature of existing studies and the critical gaps in current surveys. To unify the research landscape shaped by differing assumptions, fairness dimensions, and application contexts, we integrate multiple aspects, including bias sources, fairness definitions, technical mitigation strategies, evaluation resources, and ethical considerations, into a coherent education-centered framework. Furthermore, we introduce a novel taxonomy featuring a multi-level dimension, explicitly capturing fairness issues that simultaneously manifest at individual and group levels. Additionally, we explicitly address key practical challenges, such as censored or partially observed learning outcomes and the inherent difficulty of selecting fairness with utility, enhancing the applicability and realism of fairness assessments in real-world educational AI scenarios. Specifically, the main contributions of this survey are as follows:

\begin{itemize}

\item Our survey explicitly situates AI fairness within educational contexts, integrating technical fairness analyzes with educational theories, practical challenges, and real-world applications, thus addressing both the fragmented research landscape and the limited consideration of fairness in existing educational AI surveys.

\item We propose a novel taxonomy categorizing biases into individual-level, group-level, and the newly introduced multi-level dimension, forming a harmonized analytical framework. This taxonomy systematically consolidates bias sources, fairness definitions, technical mitigation approaches, evaluation resources, and ethical considerations into a structured and coherent synthesis tailored to educational contexts.

\item We highlight and analyze key practical challenges in educational AI, including handling censored or partially observed outcomes such as incomplete student records, addressing the inherent complexity of biases simultaneously impacting individual students and student groups, and navigating the choice between fairness and utility, such as in prediction and assessment tasks within education. By explicitly addressing these complexities, our survey ensures that fairness evaluations and frameworks are practically applicable and realistic for educational AI settings.

\item Finally, we discuss open challenges and outline future research directions, including balancing personalization with data privacy, addressing the digital divide in AI access and infrastructure, and improving teacher and student adoption through targeted training and support. These directions underscore the need for inclusive, transparent, and context-aware strategies to ensure equitable integration of AI in education.
\end{itemize}

\textbf{Survey Structure.} We begin by outlining the methodology used to conduct our literature review in \textbf{Section \ref{sec:literature}}. Then, \textbf{Section \ref{sec:casestudy_bias}} categorizes common biases in educational AI and illustrates them through real-world case studies. Based on these insights, \textbf{Section \ref{sec:notions} } and \textbf{\ref{sec:fairness_notion}} defines notions and fairness definitions tailored to educational contexts, followed by \textbf{Section \ref{sec:approaches}}, which explores bias mitigation methods categorized by fairness objectives and implementation stages. Subsequently, \textbf{Section \ref{sec:tools_datasets}} reviews analytical tools and benchmark datasets for bias evaluation.\textbf{Section \ref{sec:ethical_considerations}} discusses key ethical principles and regulatory frameworks guiding AI deployment in education, while \textbf{Section \ref{sec:challenges_fd}} highlights open challenges and future directions.

\section{Literature Selection Criteria}
\label{sec:literature}
The selection of literature for this survey follows stringent inclusion and exclusion criteria to ensure source relevance and quality, conference proceedings, and authoritative reports from the past decade related to AI-driven application in education, particularly with respect to fairness, bias, and ethics. The search spans multiple databases, including IEEE Xplore, ACM Digital Library, PubMed, Scopus, and Google Scholar, using keywords such as ``artificial intelligence in education'', ``AI fairness in education'', and ``bias in educational AI systems''. We employ a systematic review methodology that involves identifying the literature, screening texts based on set criteria, thoroughly evaluating texts for relevance and extracting pertinent data for synthesis. Multiple reviewers carry out this process to minimize selection bias, aiming to provide a comprehensive overview of fairness and bias in educational AI, synthesizing existing research to inform future advancements in the field.

\section{AI in Education: Bias Perspectives and Case Studies}
\label{sec:casestudy_bias}

\begin{figure}[htpb] 
  \centering
  \includegraphics[width=0.97\textwidth]{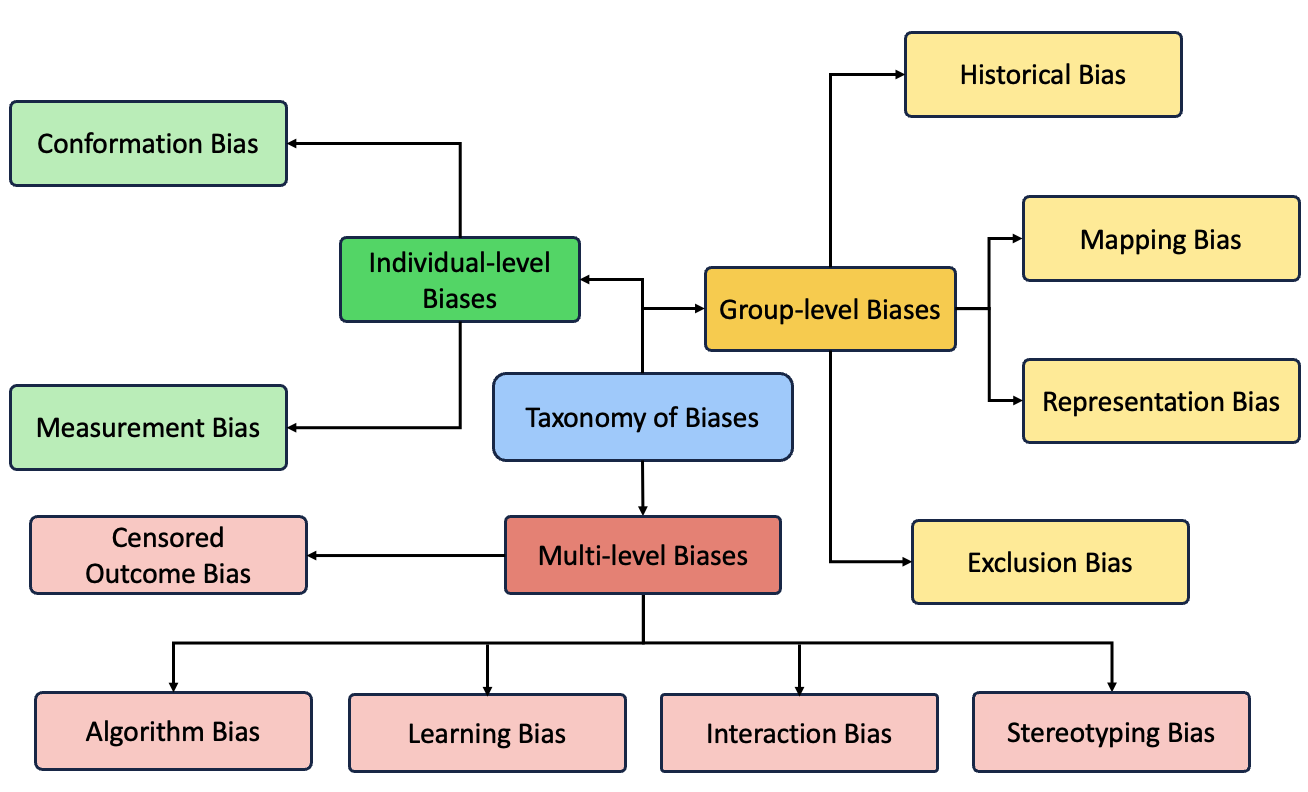} 
  \captionsetup{justification=centering} 
  \vspace{0.05cm}
  \caption{Taxonomy of biases in educational AI, categorized into group-level, individual-level, and multi-level biases.} 
  \label{fig: Types of bias} 
\end{figure}

Through our comprehensive literature review, we identified multiple biases in educational AI systems, each significantly affecting fairness in students' educational outcomes and experiences. To clearly align bias characterization with fairness metrics, we classify these biases according to the primary unit of harm within educational contexts: \textbf{(1) group-level biases}, which systematically disadvantage specific student groups based on demographic factors;  \textbf{(2) individual-level biases}, which involve unfair or inconsistent treatment of similar students; and importantly, \textbf{(3) multi-level biases}, a newly considered dimension capturing biases that simultaneously affect fairness at both individual and group levels through complex interactions between individual characteristics and broader demographic patterns (see Figure~\ref{fig: Types of bias}). The inclusion of multi-level biases addresses critical fairness dynamics overlooked by conventional classifications, highlighting scenarios where biases at one level propagate or exacerbate inequalities at another. In the following, we first present detailed real-world case studies to illustrate how these biases manifest within educational settings and subsequently provide precise definitions and analyses of each bias type as articulated by our proposed framework.

\subsection{Group-level Biases}
\label{sec:group bias}

Group-level bias in educational AI refers to systemic inequalities in educational outcomes that affect specific groups rather than individual students. These biases typically stem from historical inequities, underrepresentation in training data, or assumptions within algorithms that unfairly disadvantage or favor entire groups based on attributes such as race, gender, socioeconomic status, or linguistic background. As educational institutions increasingly rely on AI for critical decisions, such as admissions, resource allocation, or performance evaluation, addressing group-level bias becomes essential for promoting equitable education. In the following, we will highlight several common forms of group-level bias in educational AI and illustrate their real-world impacts through detailed case studies.

\begingroup
\tcbset{
  mycase/.style={
    enhanced jigsaw, breakable, sharp corners, frame hidden,
    borderline north={1pt}{0pt}{black},
    borderline south={1pt}{0pt}{black},
    boxsep=2pt,left=2pt,right=2pt,top=2.5pt,bottom=2pt,
    colback=gray!10
  }
}


\begin{tcolorbox}[mycase]
\noindent \textbf{Case Study:}
\noindent\textbf{College Admissions.} 

In certain college admissions processes across the United States, institutions have increasingly adopted AI-driven models to streamline applicant evaluations and selection decisions. These predictive models typically rely on historical admissions data, encompassing elements such as high school GPA, standardized test scores, extracurricular participation, and demographic profiles linked to specific zip codes or school districts. Upon a detailed review, these AI-driven admission tools were found to unintentionally favor applicants from historically privileged schools and affluent neighborhoods. In contrast, the models systematically disadvantaged applicants from economically disadvantaged areas or schools predominantly attended by racial and ethnic minority groups~\cite{bowers2017quantitative}. Notably, students applying from lower-income urban districts or rural communities were frequently assigned lower predicted success scores, resulting in reduced admission rates, despite their comparable academic qualifications and potential. This disparity arose directly from biases inherent in the historical data, which mirrored longstanding structural inequalities, rather than reflecting an objective assessment of individual student capabilities.

\end{tcolorbox}
\vspace{+0.01cm}

\noindent\textbf{Historical Bias:} As demonstrated by the college admissions scenario, historical bias in educational AI emerges when predictive models are trained on data reflecting past educational inequalities, leading to decisions that inadvertently reinforce outdated or discriminatory norms~\cite{ONeil2016Weapons,Buolamwini2018}. Such biases are commonly originate because historical datasets encapsulate systemic disparities in educational opportunities, resources, and outcomes. These disparities typically stem from structural inequalities, such as socioeconomic barriers, racial segregation, or discriminatory practices in resource allocation and educational policies, rather than reflecting students' actual academic capabilities or potential. When educational AI systems leverage these biased historical records, they risk perpetuating cycles of disadvantage—systematically favoring already privileged student groups while undervaluing or inaccurately assessing students from marginalized backgrounds.


\begingroup
\tcbset{
  mycase/.style={
    enhanced jigsaw, breakable, sharp corners, frame hidden,
    borderline north={1pt}{0pt}{black},
    borderline south={1pt}{0pt}{black},
    boxsep=2pt,left=2pt,right=2pt,top=2.5pt,bottom=2pt,
    colback=gray!10
  }
}
\begin{tcolorbox}[mycase] 
\noindent\textbf{Case Study: Moroccan At‑Risk Students Model.}

In the Moroccan education system study by Belhabib \textit{et al.}~\cite{belhabib2024teaching}, an AI model was built to identify students at risk of dropping out. The researchers analysis using SHAP revealed that the importance of predictor variables, such as previous academic performance, socioeconomic status, or regional indicators, varied significantly depending on the student subgroup. For instance, ``previous grades'' can be highly predictive for students in well‑resourced urban schools, but much less so in rural regions, where resource constraints or data quality issues make past grades a weaker indicator. Because the final model treated users uniformly rather than differentiating by subgroup, the risk exists that some students, particularly in underrepresented districts, might be misclassified as low risk when they are not, denying them needed support, or conversely classified as high risk based on predictors that are poor proxies in their context.

\end{tcolorbox}


\noindent\textbf{Mapping Bias:} As the above case illustrates, predictive models trained on predominantly urban school data may inaccurately generalize relationships to rural or socioeconomically disadvantaged schools, leading to inadequate predictions for underrepresented groups. In an educational AI setting, such inaccuracies exemplify mapping bias, which occurs when a single universal predictive model is applied uniformly across diverse student populations without accounting for subgroup-specific differences or individual learning contexts~\cite{Patton2020ContextualAnalysis}. Mapping bias emerges from the flawed assumption that the relationship between input features (\textit{e.g.}, student background, prior performance, learning conditions) and educational outcomes (\textit{e.g.}, grades, dropout risks) remains consistent across all students and subgroups—an assumption frequently invalidated by real-world educational diversity. Consequently, applying these universal models systematically disadvantages students whose learning trajectories or academic outcomes deviate from dominant patterns, thereby reinforcing inequities and disproportionately benefiting already-advantaged populations. Thus, mapping bias in educational AI primarily manifests as a group-level bias due to its differential impacts across student subpopulations.


\begin{tcolorbox}[boxrule=0pt,frame hidden,sharp corners,enhanced,borderline north={1pt}{0pt}{black},borderline south={1pt}{0pt}{black},boxsep=2pt,left=2pt,right=2pt,top=2.5pt,bottom=2pt]
\noindent\textbf{Case Study: ESL Students in Automated Grading.}
A targeted examination of automated essay grading systems, widely deployed across prominent e-learning platforms, revealed significant representation bias adversely affecting English as a Second Language (ESL) students~\cite{Holstein2019Improving}. Specifically, researchers identified that ESL students consistently received lower scores compared to native English-speaking students, highlighting systematic inaccuracies rooted in underrepresentation of ESL-authored essays in the models' training data. This imbalance in the dataset caused AI grading algorithms to inadequately recognize linguistic structures, vocabulary usage, and grammatical patterns common among ESL learners, thus misinterpreting their intended meanings and educational competencies. As a direct consequence of this representation bias, ESL students experienced unfair disadvantages, including decreased academic confidence, hindered educational progression, and limited opportunities for further advancement. 

\end{tcolorbox}

\noindent\textbf{Representation Bias:} As illustrated by the case study on ESL students in automated grading, insufficient representation of ESL-authored essays within training datasets led to biased evaluation models that consistently assigned lower scores to these students. Therefore, the representation bias occurs when certain demographic groups or characteristics are systematically underrepresented or misrepresented in training data, causing AI models to perpetuate existing societal biases and educational inequalities~\cite{Buolamwini2018} in educational AI setting. Therefore, representation bias primarily manifests as a group-level bias by disproportionately disadvantaging specific student groups whose characteristics and learning patterns are inadequately captured in AI training processes~\cite{Shahbazi2022RepresentationBI}. The bias consequently impaired ESL students' confidence, academic performance, and educational progression, indicating the critical necessity for representative training data in educational AI development.


\begingroup
\tcbset{
  mycase/.style={
    enhanced jigsaw, breakable, sharp corners, frame hidden,
    borderline north={1pt}{0pt}{black},
    borderline south={1pt}{0pt}{black},
    boxsep=2pt,left=2pt,right=2pt,top=2.5pt,bottom=2pt,
    colback=gray!10
  }
}
\begin{tcolorbox}[mycase]
\noindent\textbf{Case Study: Accessibility in Educational Tools.}

In widely adopted online educational tools whose interfaces and content primarily catered to native English speakers without adequately accommodating multilingual learners or students with disabilities~\cite{Holstein2019,Zheng2018,unesco2012exclusion}. Detailed evaluations showed that students from non-English speaking backgrounds frequently struggled due to insufficient multilingual support, leading to disengagement and poorer educational outcomes. Further analysis revealed that students with visual, hearing, or cognitive impairments encountered substantial accessibility barriers, as tools lacked essential assistive features such as alternative text, closed captions, or adaptive navigation tools. Consequently, these platforms unintentionally widened existing educational disparities by systematically disadvantaging marginalized groups. 


\end{tcolorbox}

\noindent\textbf{Exclusion Bias:} As highlighted in the accessibility case study, exclusion bias refers to the systematic marginalization of specific groups resulting directly from the design, operation, or implementation of AI systems. Within educational AI contexts, exclusion bias arises when user interfaces, content, or interaction modalities fail to accommodate diverse student needs, particularly affecting groups such as multilingual learners or students with disabilities. Rather than impacting students randomly or individually, exclusion bias systematically disadvantages entire demographic groups by reinforcing existing barriers related to language or accessibility. Thus, this type of bias can reinforce existing inequities and restrict opportunities for marginalized groups within educational settings.

\subsection{Individual-level Biases} 
\label{sec:individual bias}


In educational AI, individual‑level bias arises when systems treat students with comparable learning profiles or abilities differently, resulting in inconsistent or unfair outcomes at the single‑learner level. This bias can emerge from subtle flaws in training data or model design, leading to unequal grading, misclassifications of risk, or uneven personalization among otherwise similar students. Recognizing these effects is critical to ensure that AI-driven educational tools support equitable learning at the learner level. The following will highlight two common forms of individual-level bias in educational AI systems through case studies.

\begingroup
\tcbset{
  mycase/.style={
    enhanced jigsaw, breakable, sharp corners, frame hidden,
    borderline north={1pt}{0pt}{black},
    borderline south={1pt}{0pt}{black},
    boxsep=2pt,left=2pt,right=2pt,top=2.5pt,bottom=2pt,
    colback=gray!10
  }
}
\begin{tcolorbox}[mycase]
\noindent\textbf{Case Study: Automated Essay Scoring System.} 

In automated essay scoring systems (AES), which tend to assign lower scores to essays written in African American Vernacular English (AAVE) while awarding higher scores to those in Standard American English (SAE)~\cite{shermis2003automated}, measurement bias prominently manifests at the individual level. Research published in journals such as the Journal of Educational Measurement indicates that this disparity arises because AES systems primarily rely on training datasets dominated by Standard American English, causing linguistic or cultural differences at the individual level to distort assessments of a student's true abilities.
\end{tcolorbox}

\noindent\textbf{Measurement Bias:} The previously discussed case of Automated Essay Scoring (AES) systems illustrates how measurement bias can manifest when models rely predominantly on training datasets containing essays written in Standard American English (SAE). Measurement bias in educational AI arises when data collection methodologies inaccurately represent students' true abilities or knowledge due to systematic distortions in assessment inputs~\cite{Gasevic2015}. Specifically, reliance on SAE-dominated datasets leads to systematic inaccuracies and undervaluation of essays composed using alternative linguistic patterns, such as African American Vernacular English (AAVE). Consequently, students employing these alternative linguistic expressions are unfairly assessed as having lower academic proficiency or writing skills, inaccurately reflecting their genuine educational capabilities and potential. Such biases compromise the validity of educational evaluations by systematically disadvantaging students whose linguistic, cultural, or socio-economic characteristics differ from dominant norms, further exacerbating existing inequities among linguistically diverse student populations.


\begingroup
\tcbset{
  mycase/.style={
    enhanced jigsaw, breakable, sharp corners, frame hidden,
    borderline north={1pt}{0pt}{black},
    borderline south={1pt}{0pt}{black},
    boxsep=2pt,left=2pt,right=2pt,top=2.5pt,bottom=2pt,
    colback=gray!10
  }
}
\begin{tcolorbox}[mycase]
\noindent\textbf{Case Study: Bias in Personalized Learning Platforms.}

Widely-used personalized learning platforms in U.S. schools frequently exhibit confirmation bias, as they systematically prioritize educational resources and activities closely aligned with students' previous strengths, interests, or performance patterns~\cite{Selbst2019,Mannion2014SystematicBI}. For instance, students demonstrating proficiency in STEM fields predominantly receive advanced STEM-related content recommendations, rarely encountering challenging materials from arts or humanities. Conversely, students struggling initially in certain subjects are frequently presented simplified content or redirected altogether, limiting their opportunities to engage with and overcome more demanding topics. 

\end{tcolorbox}

\noindent\textbf{Conformation Bias:} Confirmation bias in educational AI refers to the algorithmic tendency to favor educational content, recommendations, or outcomes that align with students' pre-existing strengths, interests, or behaviors. As analyzed in the case study of the personalized learning platform above, confirmation bias can severely limit students' exposure to diverse subjects, novel challenges, or alternative learning strategies, thus hindering comprehensive academic development~\cite{Rader2018Explanations,eubanks2018automating}. This bias emerges from predictive models and recommendation algorithms optimizing primarily for short-term engagement, task completion, or immediate positive feedback. Confirmation bias reinforces existing skills and preferences, preventing students from broadening their academic horizons. At a systemic level, confirmation bias may reinforce educational inequalities by limiting the breadth of educational experiences for underserved student populations. Recognizing and mitigating this bias is essential for fostering equitable, inclusive, and holistic educational outcomes.

\subsection{Multi-level Biases}


We further expand the existing fairness taxonomy by introducing the concept of multi-level biases, a category we propose to capture fairness interactions that occur simultaneously across individual and group dimensions in educational AI. Multi-level biases represent interdependent fairness dynamics that arise when addressing one level of fairness inadvertently influences the other. Unlike traditional classifications that separately consider individual fairness  and group fairness, our proposed framework explicitly recognizes the intertwined nature of these dimensions. Specifically, multi-level biases encompass cases where fairness adjustments at the individual level propagate to affect demographic equity, or where group-level interventions unintentionally distort individual fairness. In educational contexts, this integrated perspective is essential, as both individual learning outcomes and collective equity objectives must be balanced to ensure truly fair AI-driven decisions. In the following sections, we provide real-world educational examples to illustrate these multi-level fairness dynamics and demonstrate the necessity of this expanded classification for developing comprehensive mitigation strategies.

\begingroup
\tcbset{
  mycase/.style={
    enhanced jigsaw, breakable, sharp corners, frame hidden,
    borderline north={1pt}{0pt}{black},
    borderline south={1pt}{0pt}{black},
    boxsep=2pt,left=2pt,right=2pt,top=2.5pt,bottom=2pt,
    colback=gray!10
  }
}
\begin{tcolorbox}[mycase]
\noindent\textbf{Case Study: Medical Education Bias via LLMs.}

Consider the use of large language models (LLMs), such as GPT, in medical education, which are frequently employed to support curriculum design, student assessments, and personalized learning pathways. These models vividly illustrate algorithmic bias operating across multiple fairness dimensions. At the individual level, due to imbalances or insufficient diversity in training datasets, these LLMs may deliver inaccurate or misleading information tailored to specific students whose backgrounds or learning paths deviate from dominant training patterns. Extending beyond individual impacts, these inaccuracies can systematically reinforce broader inequities at the group level. For instance, if the underlying training data predominantly reflect Western or male-centric medical knowledge, the models might systematically underrepresent critical insights relevant to underrepresented student populations (\textit{e.g.}, women or ethnic minorities), thereby exacerbating structural disparities in medical education outcomes and potentially impacting healthcare services for these communities. This case underscores the importance of first ensuring individual fairness, providing accurate and unbiased content tailored to each learner, and subsequently extending these fairness principles to maintain consistency and equity across diverse student groups.

\end{tcolorbox}

\noindent \textbf{Algorithm Bias: }The scenario described above clearly illustrates how algorithmic biases in educational AI systems can simultaneously span both individual and group dimensions, reinforcing and magnifying inequities across educational contexts. Unlike biases that solely produce inconsistent outcomes for individual learners or exclusively disadvantage entire demographic groups, algorithmic bias typically originates from individual-level inconsistencies and extends toward reinforcing broader group-level inequities. Such biases may result from biased training data, flawed model designs, or the broader institutional and societal contexts in which algorithms operate~\cite{Fu2020AIAA}. For instance, an algorithm might inaccurately classify or recommend learning resources for particular students based on atypical individual data patterns, consequently exacerbating systemic disparities affecting entire student populations based on race, gender, or socioeconomic status. In this way, recognizing and addressing algorithmic bias highlights the need to consider individual fairness while simultaneously ensuring equity across demographic groups.


\begingroup
\tcbset{
  mycase/.style={
    enhanced jigsaw, breakable, sharp corners, frame hidden,
    borderline north={1pt}{0pt}{black},
    borderline south={1pt}{0pt}{black},
    boxsep=2pt,left=2pt,right=2pt,top=2.5pt,bottom=2pt,
    colback=gray!10
  }
}
\begin{tcolorbox}[mycase]
\noindent\textbf{Case Study: Bias in AI-Based Early Warning Systems in K–12 Education.} 


In the United States, several K–12 districts have implemented AI-driven Early Warning Systems (EWS) designed to proactively identify students at risk of poor academic outcomes, dropout, or disciplinary issues, thereby enabling targeted interventions~\cite{Celi2022SourcesOB}. While these systems often show strong overall predictive performance, deeper analyses have revealed significant multi-level biases. Initially, biases emerged prominently at the group level, with African American and Hispanic students disproportionately labeled as "at-risk," due to historically biased disciplinary records and attendance data influenced by socioeconomic factors. This over-identification at the group level led to systemic stigmatization and misallocation of educational resources. Extending beyond these group-based biases, the effects subsequently permeated the individual level, inaccurately labeling specific minority students as high-risk solely based on demographic group patterns rather than their actual individual behaviors or academic achievements. Such misclassification resulted in inappropriate interventions, negatively affecting students' self-efficacy, motivation, and overall educational experiences. Thus, this case underscores the importance of beginning fairness analyses at the group level and then refining evaluations to ensure consistent and accurate treatment at the individual student level, promoting comprehensive fairness in educational AI systems.


\end{tcolorbox}

\vspace{0.3cm}

\noindent\textbf{Learning Bias:} As demonstrated by the previously discussed case of AI-driven Early Warning Systems (EWS), models trained to minimize prediction errors overall may unintentionally over-identify certain student groups (such as African American and Hispanic students) as “at-risk,” due to historical biases embedded within disciplinary and attendance data~\cite{Celi2022SourcesOB}. This issue exemplifies learning bias in educational AI, which occurs when modeling decisions inadvertently amplify disparities in predictive outcomes across different students or demographic groups~\cite{Bagdasaryan2019Differential}. Learning bias frequently arises from modeling choices during training—particularly the selection of objective functions that prioritize predictive accuracy or specific performance metrics. Consequently, individual students may be inaccurately labeled based on their group's historical data patterns rather than their personal academic performance, undermining targeted and effective educational interventions. Simultaneously, at the group level, learning bias exacerbates existing educational inequities by disproportionately impacting marginalized populations. Therefore, recognizing and addressing the intertwined impacts at both individual and group dimensions is critical to developing educational AI systems that ensure fairness and equitable outcomes for diverse student communities.


\begin{tcolorbox}[boxrule=0pt,frame hidden,sharp corners,enhanced,borderline north={1pt}{0pt}{black},borderline south={1pt}{0pt}{black},boxsep=2pt,left=2pt,right=2pt,top=2.5pt,bottom=2pt]
\noindent\textbf{Case Study: Adaptive Learning Platforms in Higher Education.}



An in-depth analysis of adaptive learning platforms used in higher education revealed a distinct pattern of interaction bias~\cite{Rader2018Explanations}. At the individual level, students who actively engaged with the platform benefited from highly personalized and adaptive learning experiences, while less-engaged students received limited personalization that often failed to meet their actual learning needs. Over time, this imbalance at the individual level aggregated into systemic inequities across groups. Specifically, students with lower engagement—who disproportionately came from marginalized or socioeconomically disadvantaged backgrounds, were collectively deprived of effective personalization and academic support. This dynamic created a self-reinforcing cycle, where individual-level disparities evolved into broader group-level inequalities, further widening existing achievement gaps. 


\end{tcolorbox}

\noindent\textbf{Interaction Bias:} The case study above demonstrates that highly active students tend to disproportionately influence algorithmic recommendations and adaptations, resulting in personalized experiences tailored primarily to their interactions~\cite{Farnadi2018UserProfiling}. This scenario exemplifies interaction bias in educational AI, which occurs when disparities in student engagement lead to biased outcomes, simultaneously impacting fairness at both individual and group levels. Interaction bias creates a feedback loop where inconsistent treatment among individual students emerges based on their differing engagement levels. Additionally, certain student populations, particularly those from socioeconomically disadvantaged or linguistically diverse backgrounds, often exhibit systematically lower engagement levels, thereby receiving less effective personalization and fewer learning support opportunities. Consequently, interaction bias not only affects individual student experiences but also reinforces and exacerbates educational inequalities across diverse student groups.



\begingroup
\tcbset{
  mycase/.style={
    enhanced jigsaw, breakable, sharp corners, frame hidden,
    borderline north={1pt}{0pt}{black},
    borderline south={1pt}{0pt}{black},
    boxsep=2pt,left=2pt,right=2pt,top=2.5pt,bottom=2pt,
    colback=gray!10
  }
}
\begin{tcolorbox}[mycase]
\noindent\textbf{Case Study: AI-Driven Career Guidance Systems.}


A prominent case involved an AI-driven career guidance platform widely implemented in high schools across the U.S.~\cite{Noble2018,Zhao2017MenAlsoLikeShopping}. Rigorous analyses revealed that this platform consistently recommended careers aligned closely with traditional gender stereotypes, disproportionately suggesting nursing, teaching, or humanities-related professions for female students, while directing male students predominantly toward STEM and technical fields. These group‑level disparities subsequently manifested at the individual level,  students received limited, stereotyped recommendations, unintentionally constraining their perceptions of potential career paths and perpetuating gender inequities in professional fields. This case exemplifies how group‑originated bias can propagate to individual outcomes, underscoring the importance of refining group fairness interventions to preserve equitable treatment for all learners.

\end{tcolorbox}

\noindent\textbf{Stereotyping Bias:} As demonstrated by the earlier case of the AI-driven career guidance platform, stereotyping bias can lead systems to produce recommendations aligned with entrenched gender norms, thereby reinforcing harmful stereotypes. This phenomenon exemplifies stereotyping bias in educational AI, which arises when algorithms make assumptions based on generalized societal stereotypes, leading to discriminatory or restrictive outcomes. When educational AI systems embed these societal biases into their recommendation or evaluation processes, they risk systematically perpetuating group-level inequalities by reproducing historical stereotypes within learning and career pathways~\cite{Zhao2017MenAlsoLikeShopping,Bolukbasi2016}. Beyond group-level effects, such biases can also constrain students’ personal development and aspirations, limiting their exploration of diverse academic or professional opportunities beyond socially constructed expectations.

\begingroup
\tcbset{
  mycase/.style={
    enhanced jigsaw, breakable, sharp corners, frame hidden,
    borderline north={1pt}{0pt}{black},
    borderline south={1pt}{0pt}{black},
    boxsep=2pt,left=2pt,right=2pt,top=2.5pt,bottom=2pt,
    colback=gray!10
  }
}
\begin{tcolorbox}[mycase]
\noindent\textbf{Case Study: Case Study: Bias in AI-Based Early Warning Systems.}
In a major metropolitan school district, an AI-based early warning system (EWS) was deployed to predict student dropout risk and identify those needing targeted support~\cite{chalkbeat2023dews}. However, due to censored educational data, arising when students transferred schools, took prolonged leave, or remained continuously enrolled without definitive outcomes—the system frequently encountered incomplete student records. Upon close examination, it became evident that students from economically disadvantaged backgrounds and immigrant communities disproportionately had incomplete or partially observed educational trajectories. As a result, the predictive model, trained primarily on fully observed historical data, consistently labeled these students as at higher dropout risk or academically vulnerable due solely to their incomplete data records. This issue led many students receiving inappropriate or premature interventions, simply because their educational outcomes were not yet fully known or recorded.

\end{tcolorbox}



\noindent \textbf{Censored Outcome Bias:} 
As illustrated by the previously discussed case of AI-based Early Warning Systems (EWS)~\cite{chalkbeat2023dews}, predictive models relying on incomplete educational data can disproportionately misclassify students from marginalized communities, such as economically disadvantaged and minority groups, due to their higher likelihood of having censored or partially observed outcomes. This issue exemplifies Censored Outcome Bias in educational AI, defined as systematic inaccuracies in predictions stemming from uncertainties around student outcomes such as graduation or dropout statuses—often caused by student transfers, extended enrollments, or unreported final results. Beyond its direct connection to incomplete data, Censored Outcome Bias uniquely manifests across individual, group, and multi-level fairness dimensions. At the individual level, censored data may cause significant prediction errors, leading to incorrect labels and inappropriate educational interventions, negatively affecting students' academic trajectories and self-confidence. Meanwhile, at the group level, marginalized communities, including economically disadvantaged and minority students, are disproportionately impacted, as they are more likely to experience incomplete records. Consequently, predictions for these groups systematically perpetuate inequities and harmful stereotypes. Furthermore, Censored Outcome Bias inherently exhibits multi-level characteristics, where student individual misclassifications aggregate into broader student group inequities, intensifying existing educational disparities. Addressing this complex interplay underscores the necessity for educational AI frameworks explicitly designed to manage censored outcomes comprehensively, ensuring equitable predictive outcomes for all students.

\section{Notations}
\label{sec:notions}

To rigorously formalize fairness within educational AI contexts, we first establish a consistent notation capturing key elements relevant to the definition and measuring fairness. Consider an educational dataset \(D\) designed for a binary prediction task, characterized by a target attribute \(Y = \{+, -\}\), such as pass/fail courses or at-risk/normal classifications. We assume the presence of a binary protected attribute, denoted as \(A \in \{a, \bar{a}\}\), such as socioeconomic status (\textit{e.g.}, economically disadvantaged vs. advantaged students or male and female). Within this notation, \(a\) explicitly indicates the protected or disadvantaged student group, while \(\bar{a}\) corresponds to the non-protected or advantaged group. Predictions generated by the educational AI models are represented as \(\hat{Y} = \{+, -\}\). To clearly distinguish predictions across student groups, we introduce refined notation tailored to educational settings: \(a^+\) and \(a^-\) represent students within the disadvantaged group predicted as positive (\textit{e.g.}, predicted to pass or succeed academically) or negative (\textit{e.g.}, predicted to fail or be at-risk), respectively; similarly, \(\bar{a}^+\) and \(\bar{a}^-\) represent corresponding predictions for students in the advantaged group. With this notation established, we can now formally evaluate the predictive performance of educational AI models using standard classification metrics tailored to educational outcomes, such as True Positives (TP; correctly identifying students who succeed or pass), False Negatives (FN; incorrectly identifying successful students as at-risk or failing), True Negatives (TN; correctly identifying students who do not succeed or are at-risk), and False Positives (FP; incorrectly predicting students as successful when they are actually at-risk or unsuccessful).


\section{Fairness Definitions in AI in Educational Settings}
\label{sec:fairness_notion}


As discussed in Section \ref{sec:casestudy_bias}, bias in educational AI spans individual, group, and multi-level dimensions. To systematically address these issues, the primary key lies in translating the concept of bias into operational definitions of fairness, supported by rigorous and repeatable evaluation metrics. This section establishes a definition of fairness for educational AI and establishes a connection between identified biases and quantifiable fairness standards, thereby laying the theoretical foundation for subsequent sections.

\subsection{Group Fairness} 

Group fairness addresses equity at the demographic group level, ensuring that AI-driven educational decisions and outcomes do not systematically disadvantage specific student groups based on sensitive attributes such as race, gender, linguistic background, or socioeconomic status~\cite{heidari2019moral}. In educational settings, achieving group fairness involves identifying and correcting biases, such as historical bias, representational bias, which often lead to persistent educational disparities across entire student populations. For example, in AI-based early warning systems or automated admissions processes, group fairness ensures that no specific demographic systematically experiences higher rates of adverse outcomes, such as being misclassified as ``at risk'' or receiving unequal access to advanced educational resources, thereby promoting equitable treatment and reducing systemic educational inequality~\cite{hu2020towards}. To effectively operationalize these fairness objectives, quantitative metrics are necessary to evaluate and guide the fairness performance of educational AI systems. In the following, we introduce several group fairness metrics, which enable the detection, quantification, and subsequent mitigation of unequal treatment among different student populations.

\noindent \textbf{Statistical Parity(SPD).}
In educational settings, predictive models often assist in critical decisions such as identifying students who should receive additional academic support, predicting academic success, or allocating educational resources. However, when these models are trained on historically biased or under-representative data, certain student groups (\textit{e.g.}, economically disadvantaged or minority students) might systematically receive fewer positive predictions (\textit{e.g.}, being less frequently identified as candidates for advanced classes or academic support). This can perpetuate existing educational inequalities, limiting opportunities for affected student populations. To systematically evaluate and quantify such disparities, the fairness metric known as Statistical Parity~\cite{Dwork2012Fairness} is commonly employed. Statistical Parity ensures positive predictive outcomes as $\hat{Y}$, such as predicted academic success or eligibility for interventions, are proportionally and equitably distributed across protected ($a$) and non-protected ($\bar{a}$) student groups, within an acceptable margin $\epsilon$:

\begin{equation}
|P(\hat{Y}=+|A=a) - P(\hat{Y}=+|A=\bar{a})| \leq \epsilon
\end{equation}

\noindent Specifically, the Statistical Parity Difference (SPD) explicitly quantifies this disparity:

\begin{equation}
SPD =\lvert P(\hat{Y}=+|A=a) - P(\hat{Y}=+|A=\bar{a})\rvert
\end{equation}

\noindent The value of SPD ranges from $-0$ to $1$, with $SPD=0$ indicating perfect fairness across groups, and higher SPD values signifying increased bias against the protected student group.





\noindent \textbf{Equal Opportunity (EO).} 
AI models are frequently used in the education field to recognize students with genuine potential for success or academic achievement. However, inherent biases embedded within historical performance data can cause certain groups—such as students from economically disadvantaged backgrounds or ESL students—to be systematically overlooked by predictive algorithms, thereby underestimating their true academic abilities. Such inequities may deprive these students of critical academic opportunities, affecting everything from enrolling in advanced courses to qualifying for necessary educational interventions. To specifically address systemic underestimation, Hardt \textit{et al.} introduced the important fairness metric of Equality of Opportunity (EO)~\cite{Hardt2016Equality}, emphasizing the accurate identification of genuinely qualified students across demographic boundaries. Formally, Equal Opportunity asserts that predictive models must achieve equal True Positive Rates (TPR) for protected ($a$) and non-protected ($\bar{a}$) student groups who truly belong to the positive outcome class:

\begin{equation}
P(\hat{Y} = + | A = a, Y = +) = P(\hat{Y} = + | A = \bar{a}, Y = +)
\end{equation}

\noindent Here, TPR quantifies the accuracy in correctly predicting positive outcomes:

\begin{equation}
\text{TPR} = \frac{TP}{TP + FN}
\end{equation}

\noindent Alternatively, Equal Opportunity can be expressed equivalently through the False Negative Rate (FNR):

\begin{equation}
\text{FNR} = \frac{FN}{TP + FN}
\end{equation}

\noindent Differences in EO highlight scenarios where specific demographic groups are disproportionately denied recognition for actual academic success, typically due to biased historical representations. The EO difference is computed as follows:

\begin{equation}
EO = \left| P(\hat{Y}=-|Y=+,A=a) - P(\hat{Y}=-|Y=+,A=\bar{a}) \right|
\end{equation}

\noindent The EO values range from 0, representing complete fairness, to 1, indicating maximum discrimination, thus enabling precise detection and remediation of bias in educational AI systems.

\noindent\textbf{Predictive Parity (PP).}
Even when educational AI models achieve high overall accuracy, they may still produce unreliable predictions that disproportionately affect different student groups. For example, a model might overestimate success rates for students from well-resourced schools while underestimating them for students from disadvantaged backgrounds, despite both receiving the same positive prediction. Such discrepancies undermine the credibility and fairness of AI-assisted decision-making in education. To detect and quantify these inconsistencies, the fairness measure known as Predictive Parity (PP)~\cite{Chouldechova2017} is applied. Predictive Parity ensures that the likelihood of true success among students predicted to succeed remains consistent across protected ($a$) and non-protected ($\bar{a}$) groups. Formally, the Positive Predictive Value (PPV, or precision) is given by:

\begin{equation}
\text{PPV} = \frac{TP}{TP + FP}
\end{equation}

\noindent The condition of Predictive Parity requires equal PPV across groups:

\begin{equation}
P(Y = +|\hat{Y} = +, A = a) = P(Y = +|\hat{Y} = +, A = \bar{a})
\end{equation}

\noindent The deviation from this condition is captured by the Predictive Parity difference (PP):

\begin{equation}
PP = \left| P(Y = +|\hat{Y} = +, A = a) - P(Y = +|\hat{Y} = +, A = \bar{a}) \right|
\end{equation}

\noindent Value of PP closer to zero indicate that predictions are equally reliable across student groups, while higher values reveal disparities in predictive precision—signaling that certain groups may be systematically advantaged or disadvantaged in educational predictions.








\noindent\textbf{Predictive Equality (PE).}
When such models disproportionately produce false-positive predictions—mistakenly indicating that students will succeed academically when they ultimately struggle or fail—certain demographic groups can suffer systematic harm. For instance, if a predictive model incorrectly forecasts higher academic success rates for students from particular socioeconomic backgrounds, schools may unintentionally allocate resources inequitably, disadvantaging students who genuinely require additional support. 
To specifically address this concern, Alexandra \textit{et al.} introduced the fairness metrics known as Predictive Equality (PE)~\cite{Chouldechova2017}, ensuring consistency in false positive rates across protected and non-protected student groups.  Formally, Predictive Equality requires that the probability of incorrectly predicting success for students who ultimately do not succeed remains equal across groups:

\begin{equation}
P(\hat{Y} = + | Y = -, A = a) = P(\hat{Y} = + | Y = -, A = \bar{a})
\end{equation}

\noindent The extent of deviations from Predictive Equality is measured by the absolute difference in false positive rates between groups:

\begin{equation}
PE = \left| P(\hat{Y} = + | Y = -, A = a) - P(\hat{Y} = + | Y = -, A = \bar{a}) \right|
\end{equation}

\noindent The PE ranges from 0 (indicating perfect fairness and equal predictive errors across groups) to 1 (indicating severe disparities in predictive errors). Using Predictive Equality, educational institutions can ensure that AI-driven predictions do not systematically favor or disadvantage any group, promoting equitable decision-making and more consistent treatment of students regardless of demographic background.






\noindent \textbf{Concordance Imparity (CI).} AI-driven educational predictive models frequently encounter situations where students' academic outcomes remain uncertain due to censored or incomplete data, such as when students transfer, withdraw, or continue enrollment beyond typical assessment periods. These uncertainties pose significant fairness challenges, particularly when predictions consistently disadvantage specific individuals or demographic groups by systematically underestimating their true potential or misjudging their risk status. Unlike conventional fairness metrics that assume fully observed student outcomes, Zhang \textit{et al.} propose the Concordance Imparity (CI)~\cite{zhang2021fair} explicitly addresses this challenge by assessing fairness through the predictive model's discriminative power, measured by concordance fractions, across distinct student demographic groups even under conditions of censored information. Formally, CI quantifies the maximum discrepancy in concordance fractions between student groups:

\begin{equation}
CI = \max_{g,g' \in G, g' \ne g} |F(g) - F(g')|
\end{equation}

\noindent where the concordance fraction \(F(g)\) for each student demographic group \(g\) is defined as:

\begin{equation}
F(g) = \frac{1}{|M(g)|} \sum_{i|G(x_i)=g} \sum_{j \ne i}\mathbf{1}[r(x_i) > r(x_j)]
\end{equation}

\noindent Here, the set of permissible student pairs \(M(g)\) within group \(g\) is explicitly defined as:

\begin{equation}
M(g) = \{ (x_i, x_j) \mid G(x_i) = g, x_j \ne x_i, \delta_{\min(t_i,t_j)} = 1 \}
\end{equation}

\noindent In these equations, \(r(x_i)\) denotes the predicted risk score (e.g., dropout probability) for student \(x_i\), \(G(x_i)\) indicates the demographic group of the student, and \(\delta\) represents whether the student's actual outcome has been observed (\textit{i.e.}, not censored). By leveraging Concordance Imparity, educational institutions can explicitly quantify and mitigate unfair predictive disparities resulting from censored or uncertain educational outcomes, ensuring more equitable and consistent decisions across diverse student groups.

\subsection{Individual Fairness} 

Individual fairness in educational contexts emphasizes that students with similar academic abilities and educational experiences should receive equal treatment, regardless of their demographic information. Under this fairness definition, the allocation of educational resources, opportunities, and outcomes is determined solely by relevant factors such as student performance or potential, rather than extraneous characteristics such as race, gender, or socioeconomic background~\cite{fleisher2020whats,heidari2019moral}. For example, students who demonstrate comparable abilities and learning potential should have equivalent access to educational resources, supportive experiences, and equitable outcomes, free from discriminatory influences~\cite{Dwork2012Fairness,gillen2018fairness}. Individual fairness explicitly addresses biases that arise when students are inconsistently treated based on non-educational factors, emphasizing fair and unbiased personalization and evaluation in educational AI. Compared to group fairness, the notion of individual fairness is able to enforce fairness at a finer granularity at the student level, and it does not consider any sensitive attribute. Currently, there are only a few works that consider individual fairness. We hereby present existing individual fairness metrics.

\noindent \textbf{Distance-Based Individual Fairness.} In educational AI, distance-based individual fairness emphasizes that students with similar academic features or educational profiles should receive similar predictions from predictive models, such as those used for assessing student performance, identifying at-risk students, or allocating personalized educational resources. This fairness concept is typically formalized through the Lipschitz condition, which requires that the distance between two students in the prediction (output) space should not exceed their distance in the original feature (input) space, adjusted by a scalar factor. Formally, this condition is defined as:

\begin{equation}
     d_Y(f(x_i), f(x_j)) \leq L \cdot d_X(x_i, x_j)
\end{equation}

\noindent where $f(\cdot)$ denotes the results of the predictive model for individual students and $d_Y(\cdot)$ and $d_X(\cdot)$ represent the distance metrics in the prediction (output) and educational feature (input) spaces respectively, $L$ is the Lipschitz constant that rescales the input distance between $x_i$ and $x_j$.

In educational applications, this metric is applied to educational prediction tasks, such as student performance forecasting, early-warning systems, and personalized feedback models, to verify whether similar students are treated equitably. To measure fairness under the Lipschitz condition, existing studies compute the average distance between each student’s prediction and those of their $k$-nearest neighbors in the input space. A smaller output variance among similar students indicates higher fairness, whereas a larger variance reveals inconsistencies that may disadvantage certain learners. This can be expressed as follows:

\begin{equation}
    1 - \frac{1}{n \cdot k} \left| \sum_{i=1}^{n} \hat{y}_i - \sum_{j = 1}^k\hat{y}_j \right|
\end{equation}

\noindent where $\hat{y}_i$ represents the probabilistic prediction for the student $x_i$, and $x_j$ denotes one of the $k$ most similar students to $x_i$.  $n$ is the total number of students in the dataset.

\noindent \textbf{Ranking-Based Individual Fairness.} In the context of educational AI, ensuring individual fairness typically involves confirming that students with similar characteristics or academic profiles receive similar predictive outcomes. Traditional approaches, such as the Lipschitz condition, rely on absolute distances between input features (\textit{e.g.}, academic records, demographics) and output predictions (\textit{e.g.}, academic success likelihood). However, these methods often face practical challenges, including determining appropriate scales or constants, and may not adequately reflect realistic student variability or nuanced relationships between students.

To overcome these issues, recent studies have proposed adopting a ranking-based perspective on individual fairness. This approach evaluates fairness by comparing two ranked lists for each student: one derived from pairwise similarities based on students' input features and the other based on similarities in predicted educational outcomes. Ideally, perfect individual fairness would imply that these two rankings align exactly for each student. In practice, fairness is commonly assessed by measuring the average similarity among the top-ranked elements across these rankings. Standard metrics employed for this purpose include Normalized Discounted Cumulative Gain (NDCG@$k$)~\cite{jarvelin2002cumulated}, introduced by Järvelin and Kekäläinen, and Chapelle \textit{\textit{et al.}} proposed Expected Reciprocal Rank (ERR@$k$)~\cite{chapelle2009expected}, both of which have been applied in educational AI fairness evaluations to assess the consistency between student similarity rankings and predicted outcomes.

Traditional ranking-based fairness metrics, while effective in evaluating the consistency between student similarities and predicted outcomes, often fail to account for incomplete or censored educational data. In many real-world educational AI applications, such as dropout prediction, longitudinal academic assessment, or performance forecasting, student outcomes may remain partially unknown due to ongoing enrollment, early withdrawal, or incomplete records. These censored observations introduce uncertainty into fairness evaluation: ranking-based models may inadvertently disadvantage students with incomplete data, leading to biased assessments that do not reflect their true academic potential.
To address this challenge, Zhang \textit{\textit{et al.}}~\cite{zhang2023individual} proposed Discounted Cumulative Fairness (DCF), an individual fairness metric built upon the ranking-based fairness framework and explicitly designed to handle censored educational outcomes. DCF extends the traditional ranking perspective by integrating censorship information directly into the fairness evaluation process. In essence, it quantifies how well the ranking of predicted outcomes (output space) aligns with the ranking of student similarities (input space), while appropriately discounting the uncertainty introduced by censored data. This enables fairer assessments of students whose outcomes are only partially observed, ensuring that missing or ongoing records do not skew fairness evaluations. Formally, DCF measures individual fairness by computing the ratio of discounted cumulative similarities from rankings derived from censored predictions relative to input-based rankings:

\begin{equation}
\text{DCF}@k = \frac{1}{N}\sum_{i=1}^{N}\frac{\text{DCG}_{R_2}(x_i)}{\text{DCG}_{R_1}(x_i)}
\end{equation}

\noindent where the discounted cumulative gain (DCG) for a student $x_i$ is calculated as:
\begin{equation}
\text{DCG}_{R}(x_i) = \sum_{pos=1}^{k}\frac{S(x_i, x_{pos})}{\log_2(pos+1)}
\end{equation}

\noindent where $k$ denotes the number of top-ranked neighboring students considered, and $S(x_i, x_{pos})$ represents the similarity between student $x_i$ and their $pos$-th neighbor. By introducing this censorship-aware adjustment, DCF offers a more reliable and realistic evaluation of fairness in educational AI systems, particularly in longitudinal or incomplete learning datasets where traditional fairness metrics fall short.

\subsection{Multi-level Fairness}

Unlike traditional fairness, multi-level fairness definitions simultaneously consider individual fairness and group fairness, explicitly addressing their interdependencies. Traditional approaches typically treat individual and group fairness independently, overlooking the possibility that achieving fairness at one level might unintentionally introduce biases at the other. Multi-level fairness addresses this issue by evaluating how individual fairness constraints are equitably distributed across demographic groups. Specifically, multi-level fairness can be viewed from two complementary directions: approaches that start with individual fairness and extend toward ensuring group consistency, and those that begin with group fairness and then refine it to maintain fairness at the individual level.

\textbf{Individual-to-Group Fairness.} Approaches within this category first emphasize fairness at the individual level, where similar individuals, such as students with comparable academic achievements, behavioral patterns, or educational contexts, should receive similar outcomes. From this individual fairness baseline, these frameworks then assess whether such fairness is consistently maintained across different demographic groups, including gender, ethnicity, or socioeconomic status. For example, in educational scenarios like admission processes or grade predictions, individual fairness implies that two students with similar academic records should obtain similar predicted outcomes. Multi-level fairness in this direction further investigates whether the fairness experienced individually is uniformly distributed across demographic subgroups. This ensures that fairness experienced by students in one group does not systematically differ from students in another, avoiding unintended subgroup disparities. For instance, Wang \textit{\textit{et al.}}~\cite{wang2024individual} proposed the concept of Individual Group Disparity (IGD), designed to quantify the consistency of individual fairness across demographic subgroups. Specifically, the IGD score measures the largest relative deviation between the group-level fairness scores (GF) across all demographic groups, effectively capturing how equitably individual fairness is applied across these groups. Mathematically, IGD is defined as follows:

\begin{align}
    \mathrm{IGD} &= \min_{\forall \ \mathbf{D}_{s_i}, \mathbf{D}_{s_j} \in \mathbf{D}, \mathbf{D}_{s_i} \neq \mathbf{D}_{s_j}} \left\{ \left| \frac{\text{Min}(\mathrm{GF}_{\mathbf{D}_{s_i}},\mathrm{GF}_{\mathbf{D}_{s_j}})}{ {\text{Max}(\mathrm{GF}_{\mathbf{D}_{s_i}},\mathrm{GF}_{\mathbf{D}_{s_j}})}} \right|  \right\}  
\end{align}

\noindent where $\mathrm{GF}_{\mathbf{D}{s_i}}$ denotes the group-level fairness score for subgroup $\mathbf{D}_{s_i}$, calculated by averaging the individual fairness ($\mathrm{IF}$) scores of each individual within the subgroup. Specifically, group-level fairness is computed as follows:

\begin{equation}
\mathrm{GF}_{\mathbf{D}{s_i}} = \frac{1}{|\mathbf{D}{s_i}|}\sum{x_j \in \mathbf{D}_{s_i}} \mathrm{IF}(x_j),
\end{equation}

\noindent where $\mathrm{IF}(x_j)$ represents the individual fairness measurement for an individual $x_j$, and $|\mathbf{D}{s_i}|$ indicates the number of individuals in subgroup $\mathbf{D}{s_i}$. By comparing group-level fairness scores across all subgroups, IGD identifies and enables mitigation of disparities in how individual fairness constraints are applied across demographics, thus ensuring equitable treatment within educational contexts.



\textbf{Group-to-Individual Fairness.} Conversely, frameworks adopting a group-to-individual fairness approach first emphasize constraints at the group level, aiming to ensure that demographic subgroups receive equitable treatment as a whole. After achieving group-level parity, such as equal acceptance rates or balanced scholarship distributions among demographic groups, these frameworks subsequently refine their constraints to prevent unintended unfairness at the individual level. Within educational contexts, this means that while demographic groups (for instance, male and female applicants) achieve comparable overall admission outcomes, additional steps are taken to ensure individual students within each subgroup are not subjected to disproportionate inequities resulting from strict adherence to these group-level constraints. The objective is thus to avoid situations where certain individuals carry an excessive burden to fulfill group fairness requirements. For instance, Wang \textit{\textit{et al.}}~\cite{wang2024group} introduced the constraint concordance difference (CCD) metric, which explicitly measures the impact of group fairness constraints on individual fairness. The CCD evaluates changes in individual fairness by comparing the concordance index of individual samples before and after applying group fairness constraints. Mathematically, this can be expressed as:

\begin{equation}
\label{equ:CCD}
CCD = R'{x_i} - R{x_i}
\end{equation}

\noindent where $R_{x_i}$ is the ranking consistency score of sample $x_i$ prior to applying group fairness constraints, and $R'_{x_i}$ is the score of the same sample after the constraints have been enforced. A key strength of CCD is its ability to pinpoint students who are disproportionately impacted when group fairness measures are strictly enforced. A lower CCD value indicates minimal individual fairness loss, signifying that fairness constraints have been effectively balanced between group equity and individual fairness. This balance is especially critical in educational settings, where AI-based decisions significantly impact individual students, including admissions processes, scholarship distributions, course placement decisions, and personalized learning recommendations. By incorporating these considerations, CCD offers a robust and reliable measure for evaluating and maintaining fairness in educational AI systems.


Although individual-to-group and group-to-individual fairness approaches begin from opposite ends of the fairness definitions, they both aim to create balanced and comprehensive fairness definitions. Individual-to-group fairness emphasizes personalized equitable treatment, making it particularly suitable for fine-grained educational interventions, such as individualized student support, personalized recommendation systems, or targeted retention efforts. Conversely, group-to-individual fairness is more aligned with educational policies aimed at institutional equity, such as standardized assessments, equitable admission policies, or fair resource distribution across different demographic groups. Overall, multi-level fairness definitions encourage educational AI systems to systematically consider fairness at both the individual and group levels, ensuring that personalized decisions support broader demographic equity and that group level decisions do not inadvertently compromise individual fairness.



\subsection{Unified Fairness-Utility Protocol}

As discussed previously, a critical practical challenge in deploying educational AI systems is the absence of unified protocols for determining appropriate selections between fairness and predictive utility. Practitioners often struggle to quantify and manage trade-offs between equitable student outcomes and predictive accuracy, particularly when predicting sensitive educational outcomes such as student performance, dropout risk, or academic achievement. For example, a model optimized solely for high predictive accuracy in identifying at-risk students might disproportionately disadvantage minority student groups, while a model focusing strictly on fairness may significantly reduce prediction accuracy and reliability. To systematically address this challenge, Wang \textit{et al.}~\cite{wang2023preventing} introduced the Fairness Bonded Utility (FBU) framework, providing a structured and unified methodology to evaluate and simultaneously balance fairness and predictive performance in educational AI applications.

The FBU framework systematically categorizes fairness mitigation techniques based on their combined effects on fairness and predictive accuracy, organizing outcomes into five distinct effectiveness levels. These levels range from simultaneously enhancing both fairness and performance to negatively impacting both metrics, clearly delineating trade-offs within a two-dimensional coordinate system. Specifically, effectiveness levels are defined as (see Figure~\ref{fig:scatter_plot} (b)):

\noindent\textbf{Level 1 (Jointly Advantageous):} Interventions that simultaneously improve fairness and predictive accuracy compared to the trade-off baseline.

\noindent\textbf{Level 2 (Impressive):} Interventions improving either fairness or predictive performance, resulting in overall superior outcomes relative to baseline.

\noindent\textbf{Level 3 (Reversed):} Interventions improving predictive accuracy but at the expense of fairness.

\noindent\textbf{Level 4 (Deficient):} Interventions that decrease either predictive accuracy or fairness, leading to overall inferior outcomes compared to baseline.

\noindent\textbf{Level 5 (Jointly Disadvantageous):} Interventions negatively impacting both fairness and predictive accuracy, representing the least favorable scenario.

Within this framework, the trade-off between fairness and predictive performance is visually represented in a two-dimensional coordinate system, as depicted in Figure~\ref{fig:scatter_plot} (a). This visualization is achieved by generating pseudo-models \(N_{\beta}\), where the parameter \(\beta\) indicates the proportion (ranging from 10\% to 100\%) of original predictions replaced by a uniform prediction label. In educational scenarios, this method simulates fairness interventions such as uniformly predicting academic success to eliminate bias, even though such interventions can potentially decrease overall prediction accuracy. An essential component of the FBU framework is the trade-off baseline, established through zero normalization, which assigns identical predictions to all students to minimize bias without regard to predictive accuracy. By incrementally increasing the proportion of uniform predictions, the baseline effectively illustrates the evolving trade-offs between fairness and performance.

\begin{figure}[h!]
  \centering
    \includegraphics[width=\textwidth]{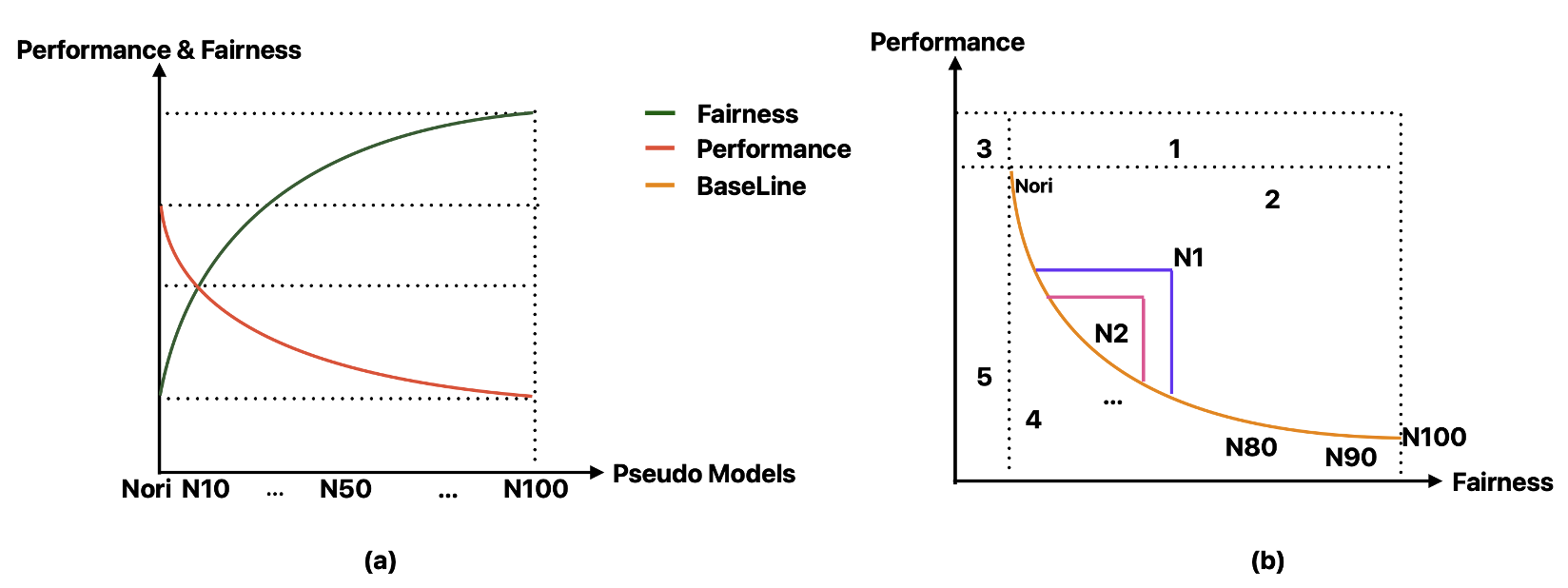}
  \caption{Visualizations of the Fairness Bonded Utility framework.}
  \label{fig:scatter_plot}
\end{figure}

Practically, FBU evaluates fairness interventions by quantifying their effectiveness relative to the baseline, primarily within the ``impressive'' region, where desirable fairness-performance balances exist. Larger enclosed areas signify more favorable trade-offs between fairness and predictive accuracy, thus enabling consistent, objective comparisons across different fairness strategies in educational contexts. Consequently, the FBU framework addresses the critical absence of unified evaluation protocols by providing a clear, structured approach to quantify and manage fairness-utility trade-offs, empowering educators, researchers, and practitioners to make informed and equitable decisions when deploying educational AI models.

\section{ Approaches to Mitigating Biases }
\label{sec:approaches}

To systematically address the aforementioned problems and biases in educational AI, researchers and practitioners have proposed various mitigation approaches. We first categorize mitigation strategies into three levels based on primary fairness goals: individual level, group level, and multi-level. Within each category, and further structure these methods according to their specific implementation stages, providing a clear and practical overview of how biases can be effectively addressed in different aspects of educational AI systems. This structured approach not only highlights the connections between fairness objectives and technical implementation, but also provides practical guidance for researchers and practitioners, facilitating precise selection and application of appropriate fairness interventions tailored to the unique requirements of educational environments. Figure~\ref{fig:Algofairness} illustrates this comprehensive organization of bias mitigation approaches.

\begin{figure}[htpb] 
  \centering
  \includegraphics[width=1\textwidth]{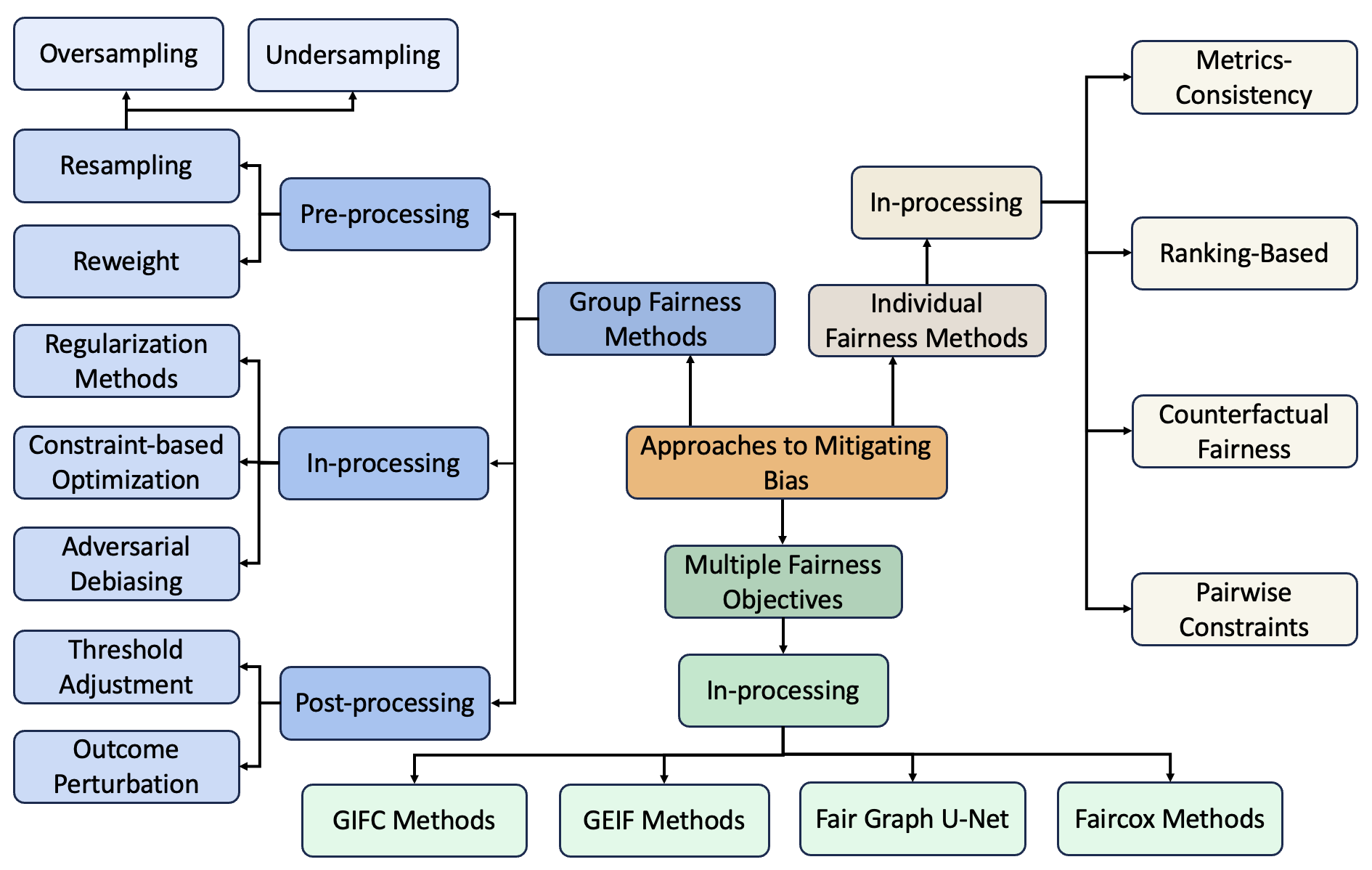} 
  \captionsetup{justification=centering} 
  \vspace{0.2cm}
  \caption{Overview of bias mitigation techniques for enhancing algorithmic fairness in educational AI, categorized into group fairness, individual fairness, and multiple fairness approaches.} 
  \label{fig:Algofairness} 
\end{figure}

\subsection{Group Fairness Methods} 


Group fairness methods aim to identify and reduce biases that systematically disadvantage particular demographic groups in educational AI systems~\cite{ferrara2023fairness}. These approaches are especially important in educational settings, where historical and societal inequities embedded in the data or model design can lead to disparate outcomes across student populations. The following subsections introduce key methods designed to promote group fairness.

\subsubsection{Pre-processing}

Pre-processing methods for group fairness focus on identifying and reducing demographic biases in educational datasets before they are used to train AI models~\cite{ferrara2023fairness}. Because educational data frequently encode historical and societal inequities, these approaches play a critical role in preventing models from inheriting systemic disadvantages that disproportionately affect certain student groups. The following subsections introduce key pre-processing strategies designed to promote group fairness in educational AI systems.


\textbf{Reweighing: } Reweighing is a crucial strategy employed in an educational setting to systematically address and mitigate biases inherent in historical or unbalanced datasets. Specifically, this technique involves adjusting the relative weights of the student instances in the dataset to ensure equitable representation and influence of historically underrepresented or marginalized student groups (such as those defined by race, gender, socioeconomic status, or linguistic background). By increasing the weights of samples from disadvantaged groups and potentially reducing the weights from dominant groups, reweighing seeks to correct historical inequities embedded in educational data. This approach ensures that predictive models, used for applications such as academic achievement forecasting, dropout risk identification, or resource allocation, accurately reflect diverse student populations, thus minimizing discriminatory outcomes. Calmon \textit{et al.}~\cite{calmon2017Optimized} introduced an optimized reweighing method specifically designed to address biases within educational datasets by treating fairness as an explicit optimization objective. This strategy systematically adjusts the weights of underrepresented student groups, effectively reducing biases in critical educational outcomes, such as dropout predictions, admissions, or resource allocations, while preserving overall predictive accuracy and utility. Such optimization-based reweighing approaches are particularly beneficial in educational contexts, where the need to maintain precise and reliable predictive performance must be carefully balanced against the imperative to ensure equitable treatment across diverse student populations. Similarly, Kamiran and Calders~\cite{kamiran2012DataPreprocessing} demonstrated the practical effectiveness of reweighing techniques to mitigate biases embedded in educational data, particularly historical biases affecting student performance predictions and resource allocations. They introduced a systematic approach in which instance weights within the training data are adjusted, increasing the influence of underrepresented or marginalized student groups such as students from lower socioeconomic backgrounds or minority groups, while decreasing the emphasis on dominant groups. This targeted reweighting reduces the impact of embedded biases by recalibrating the representation of disadvantaged groups, ensuring models reflect a fairer distribution of educational outcomes. Building upon this approach, Hajian \textit{et al.}~\cite{hajian2012methodology} proposed a comprehensive methodology specifically aimed at addressing direct and indirect discrimination within predictive models applied in educational contexts. Their method systematically analyzes educational datasets and predictive outcomes to detect biases related to sensitive demographic attributes, such as socioeconomic status, gender, or ethnicity, which could unfairly influence academic predictions or resource allocations. Once identified, these biases are proactively mitigated by adjusting the representation and influence of sensitive attributes within the model's training process, ensuring that educational predictions and subsequent interventions, such as academic tracking or resource distribution—are equitable. Consequently, this approach supports the creation of inclusive educational environments, reduces systemic disparities, and promotes fairer outcomes for diverse student populations.

\textbf{Resampling:} In the context of educational AI applications, resampling approaches play a critical role in addressing dataset imbalances and ensuring equitable representation across diverse student groups. The primary goal of these techniques in educational AI is to enhance dataset balance by either \textbf{oversampling} student groups that are underrepresented, such as minority or economically disadvantaged students, or \textbf{undersampling} dominant groups, thus mitigating biases that could negatively impact predictions of student success, dropout risk, or resource allocation. Resampling is particularly important in educational contexts such as predicting student outcomes, assessing admission fairness, or identifying at-risk students, as skewed datasets can lead to biased predictions and unequal opportunities. For instance, Kamiran and Calders~\cite{kamiran2012DataPreprocessing} demonstrated the efficacy of resampling in reducing bias in educational data mining scenarios. Similarly, Wang and Zhu~\cite{Wang2020Application} employed resampling methods to improve accuracy and fairness in predicting student achievement within digital learning environments. In general, by systematically adjusting the representation of student groups, resampling techniques help ensure that AI-driven educational decisions are fair, inclusive, and equitable, thereby supporting equal opportunities for all students. In the following, we discuss oversampling and undersampling in greater detail:

\textbf{i) Oversampling:} Oversampling techniques, such as the Synthetic Minority Oversampling Technique (SMOTE) and Generative Adversarial Networks (GANs), play a pivotal role in addressing class imbalances that arise from demographic disparities within student data. Initially developed by Chawla \textit{et al.}~\cite{chawla2002SMOTE}, SMOTE generates synthetic student data for minority groups by interpolating between existing student instances, thereby enhancing dataset representativeness. This process is particularly important in educational contexts, such as predicting dropout risk, academic performance forecasting, or personalized learning interventions, where historically marginalized student populations are often underrepresented. GANs oversampling methods, introduced by Goodfellow \textit{et al.}~\cite{Goodfellow2014GenerativeAN}, leverage adversarial training to create more nuanced and realistic synthetic student data, capturing the underlying distributions of minority groups more effectively than traditional oversampling methods. By generating authentic and representative data, GANs enable predictive models to more fairly and accurately reflect demographics of the diverse students. These oversampling approaches have shown notable effectiveness in educational applications. For example, applying SMOTE or GAN-based methods to predict academic achievement or identify at-risk students ensures predictive models consider diverse student experiences and characteristics, thus reducing biases that disproportionately affect minority groups~\cite{Fernandez2018SMOTEFL}. Through careful implementation and rigorous validation, these oversampling techniques significantly enhance the fairness, inclusiveness, and accuracy of educational AI models, ultimately supporting equitable educational outcomes across all demographics of students.


\textbf{ii) Undersampling:} While oversampling methods enrich datasets by generating additional samples, undersampling techniques such as Random Under Sampling (RUS) and the NearMiss algorithm address class imbalances by strategically reducing the size of majority classes~\cite{Liu2009ROleOfUndersampling,zhang2003NearMiss}. Specifically, these techniques strategically reduce the prevalence of majority-class student instances within training datasets to avoid biased predictions that disproportionately favor dominant demographic groups. RUS achieves dataset balance by randomly eliminating instances from overrepresented student groups, effectively mitigating biases that could skew model predictions toward majority demographics. In contrast, NearMiss strategically selects majority-class instances most similar to minority-group students based on educationally relevant metrics (\textit{e.g.}, academic performance, dropout risk), preserving more informative and representative samples within the reduced dataset. These methods have been successfully applied in educational predictive tasks such as dropout detection and identification of academically at-risk students, contributing to fairer and more equitable outcomes for historically disadvantaged or underrepresented student populations. However, undersampling also introduces notable challenges, primarily the potential loss of valuable information when instances from the majority class are discarded. This data reduction may lead to models that, although more balanced, insufficiently capture the full range of student characteristics and performance nuances, thus compromising both predictive accuracy and fairness~\cite{Haixiang2017learning,Chawla2011Editorial}. Consequently, educators and data scientists require carefully weigh the trade-offs between fairness and information retention when employing undersampling strategies. To achieve optimal outcomes, it is essential to ensure that the resulting models maintain detailed insights into student behavior and accurately reflect the complexity of educational environments.


\subsubsection{In-processing}

In-processing techniques embed group fairness constraints directly into the learning algorithms themselves, ensuring that predictive models treat different demographic groups equitably~\cite{ferrara2023fairness}. Rather than modifying the input data, these methods adjust the training objective, through fairness-aware regularization, constraint-based optimization, or adversarial debiasing, to mitigate group disparities as the model learns. We next detail representative in-processing techniques and their mechanisms for achieving group fairness in educational contexts.


\textbf{Regularization Methods:} Regularization methods in educational AI aim to enforce group-level fairness by explicitly integrating fairness considerations into the training process of predictive models. This integration typically involves adding fairness-specific penalty terms to the model's loss function, penalizing predictions that deviate significantly from fairness criteria across protected student groups. For example, Kamishima \textit{et al.}~\cite{Kamishima2011Fairness} introduced a fairness-aware regularizer designed to reduce discrimination arising from protected attributes such as gender, race, or socioeconomic status. Their method functions by adding a fairness penalty to the loss function, thereby encouraging the model to generate more equitable predictions, for instance, ensuring that predictions of academic performance or dropout risk are unbiased across demographic groups. Further extending this concept, Mary \textit{et al.}~\cite{Mary2019MultiFaceted} proposed a multi-faceted regularization strategy that simultaneously embeds multiple fairness constraints, such as equal opportunity and demographic parity, into educational predictive models. By jointly enforcing several fairness measures, their method systematically addresses different dimensions of bias, ensuring that the predictive models not only achieve robust accuracy but also equitably serve diverse student populations. Such regularization methods represent a critical mechanism for ensuring both predictive fairness and accuracy within educational AI, ultimately supporting fairer resource allocation, student performance evaluation, and intervention planning in educational settings.

\textbf{Constraint-based Optimization:} In educational data analysis, constraint-based optimization represents a distinct and powerful class of group fairness approaches, differing from regularization methods in its mechanism for enforcing fairness. Unlike regularization methods, which embed fairness considerations through penalty terms added directly to the model's loss function, constraint-based optimization incorporates explicit fairness constraints into the optimization problem itself. These methods impose specific conditions that predictive models must satisfy concerning sensitive demographic attributes, such as gender, ethnicity, or socioeconomic status. This ensures that demographic groups are treated equitably, especially in critical educational predictions such as student performance evaluation, resource allocation, or admission decisions. By embedding fairness as explicit optimization constraints, rather than relying solely on penalty-based regularization terms, constraint-based methods provide more direct and stringent guarantees of fairness compliance, making them particularly suitable for scenarios where strict adherence to demographic fairness criteria is essential. Zafar \textit{et al.}~\cite{zafar2017FairnessConstraints} introduced a pioneering framework that incorporates fairness constraints into convex optimization formulations, enabling predictive models to satisfy fairness criteria with minimal compromise to accuracy, thus preventing educational AI systems from perpetuating existing disparities among student populations. Building on this foundation, Donini \textit{et al.}~\cite{Donini2018Empirical} extended the paradigm by developing a kernel-based optimization approach capable of simultaneously accommodating multiple fairness constraints. This advancement significantly improves the adaptability and applicability of constraint-based optimization, allowing researchers and practitioners to address complex and multifaceted fairness challenges inherent in predictive educational analytics. The practical potential of constraint-based optimization has already been demonstrated in a range of educational applications. For example, Grgić-Hlača \textit{et al.}~\cite{GrgicHlaca2018BeyondDistributive} successfully applied this methodology to mitigate bias in automated essay scoring systems, promoting equitable evaluation of students from diverse linguistic and cultural backgrounds. Similarly, Saleiro \textit{et al.}~\cite{Saleiro2019Aequitas} leveraged these principles to develop a fairness-aware early-warning system for identifying students at risk of dropout, illustrating how constraint-based approaches can directly foster equitable interventions and resource allocation in real-world educational settings. Taken together, these studies underscore the versatility and effectiveness of constraint-based optimization as a means of operationalizing group-level fairness in educational AI, offering a robust framework for mitigating biases and ensuring more just outcomes across diverse student groups.

\textbf{Adversarial Debiasing:}
Adversarial debiasing represents a significant group-fairness approach in educational AI, explicitly designed to reduce predictive biases across diverse student demographics. This methodology employs a dual-model adversarial framework comprising two competing neural networks: a primary prediction model and an adversarial model. In educational contexts, the primary model aims to accurately predict critical student outcomes, such as academic achievement, dropout risk, or admission eligibility, based on available student data. Concurrently, the adversarial model attempts to infer sensitive demographic attributes (\textit{e.g.}, race, gender, or socioeconomic status) from the primary model's predictive outputs. Through iterative adversarial training, the primary model actively learns data representations that obscure demographic-specific information, making it increasingly challenging for the adversarial model to accurately recover sensitive attributes. Consequently, this training process helps ensure that the educational predictions are not unfairly biased towards or against specific student groups. To demonstrate the practical applicability and effectiveness of adversarial debiasing in real-world educational AI scenarios, Wang \textit{et al.}~\cite{wang2023preventing} successfully demonstrated the practical effectiveness of adversarial debiasing in educational AI by applying it to predictive models assessing student achievement, significantly reducing racial disparities in predicted outcomes. Similarly, Smith and Neff~\cite{smith2019GenderBias} leveraged adversarial debiasing to mitigate gender bias within online course recommendation systems, resulting in more equitable content suggestions across gender groups. Collectively, these educational applications illustrate adversarial debiasing’s capability to systematically reduce group-based biases in predictive models, thus promoting fairness, inclusivity, and equity in educational decision-making processes.


\subsubsection{Post-Processing} 


Unlike pre-processing methods that modify data inputs or in-processing strategies that adjust the learning algorithms themselves, post-processing methods aim to improve fairness by directly adjusting predictions after model training is complete~\cite{ferrara2023fairness}. In educational contexts, post-processing is particularly valuable when modifications to training data or algorithm structures are impractical or insufficient. By refining the outputs of existing educational AI models, these methods help ensure that automated assessments, recommendations, and resource allocations do not perpetuate systemic biases against any student group. This section introduces several prominent post-processing approaches specifically designed to enhance fairness in AI-driven educational applications.


\textbf{Threshold Adjustment:} Threshold adjustment represents a critical group fairness strategy in educational AI, explicitly designed to achieve equitable decision-making by strategically tuning model decision boundaries. In educational settings, predictive models frequently generate scores or probabilities—such as students' likelihood of academic success, dropout risk, or eligibility for specialized programs. The key idea behind threshold adjustment is to carefully modify these decision thresholds across demographic groups, thereby ensuring fairness objectives like demographic parity or equal opportunity are consistently satisfied. For example, a model predicting course passing rates may initially yield systematically lower success probabilities for disadvantaged student groups due to historical biases in training data. By selectively lowering the decision threshold for these groups, educational institutions can ensure equitable identification of students qualified for academic interventions or advancement, effectively mitigating group disparities. Stark \textit{et al.}~\cite{Stark2019} demonstrated threshold adjustment by embedding explicit fairness constraints directly within ranking algorithms, providing a robust methodology for educational applications involving student rankings, such as scholarship distribution or competitive admissions. Friedler \textit{et al.}~\cite{friedler2016possibility} further expanded on this approach by proposing an optimization-based framework specifically for adjusting decision thresholds to meet fairness criteria such as equal opportunity. This method ensures a fair assessment of student potential and academic achievement by systematically searching for thresholds that simultaneously satisfy both model accuracy and fairness constraints. These approaches are based on the foundational principles established by Hardt \textit{et al.}~\cite{Hardt2016}, who emphasized the importance of strategically calibrating decision thresholds to effectively enhance fairness in predictive models. Within educational AI, such meticulous threshold calibration is particularly valuable for tasks like student performance assessments, dropout predictions, admissions, and scholarship allocations. By carefully adjusting model thresholds, educational institutions can significantly reduce group-level biases, ensuring equitable access to resources and opportunities for all students. Taken together, these studies underscore that meticulous threshold calibration plays a pivotal role in promoting group fairness in educational AI, safeguarding equitable treatment of diverse student populations without substantially compromising model utility.

\textbf{Outcome Perturbation: }Outcome Perturbation represents another critical approach to achieving group fairness in educational AI. This method intentionally introduces noise or systematic adjustments into model predictions to reduce biases that disproportionately affect specific student demographic groups. Johndrow \textit{et al.}~\cite{Johndrow2017AnAF} initially developed outcome perturbation algorithms to ensure race-neutral predictions in recidivism risk models. Extending these principles to educational settings, such as dropout prediction or performance assessment, outcome perturbation can strategically obscure sensitive demographic information within model outputs, preventing unfair disadvantages to particular student groups. Feldman \textit{et al.}~\cite{Feldman2015} further advanced this methodology by systematically quantifying biases and applying targeted perturbations to certify and mitigate disparate impacts. By carefully adjusting educational predictions, such as academic performance classifications or resource allocation recommendations, this approach helps ensure demographic equity, safeguarding fair treatment for all student groups while maintaining data confidentiality through mechanisms such as differential privacy. In general, outcome perturbation methods provide an effective means of upholding fairness in educational AI, mitigating systemic bias while preserving the integrity and privacy of individual student information.

\subsection{Individual Fairness Methods}


Individual fairness methods explicitly focus on ensuring consistency and equity in the treatment of similar students, independent of their demographic information. Unlike group fairness methods, most existing individual fairness approaches rely predominantly on \textbf{in-processing} techniques, embedding fairness constraints directly within the model’s training process. This reliance arises from the fundamental requirement of individual fairness, that the predictive model must directly encode and preserve task-specific similarities among students during its learning phase. Therefore, fairness conditions require explicit implementation during model training, specialized regularization, or pairwise constraints, continuously guiding the model to produce consistent outcomes for comparable individuals. This section highlights several representative individual fairness techniques that leverage in-processing strategies to achieve unbiased and equitable educational outcomes at the individual student level.

\textbf{Metrics-consistency:} Metric-Consistency is an individual-fairness approach designed to guarantee that students with similar educational characteristics or task-relevant features receive consistent predictive outcomes from educational AI models, independent of their sensitive demographic attributes. Rooted in the foundational principle of "fairness through awareness," the core concept emphasizes that similarly situated individuals should be treated similarly based on a carefully defined notion of task-specific similarity. Unlike approaches relying purely on demographic parity or equal outcomes, metric-consistency explicitly ensures fairness by stabilizing predictions within small, meaningful neighborhoods of students.

In practice, metric-consistency is typically enforced through specialized regularization strategies integrated directly into the training process. A prevalent technique is neighborhood-based consistency regularization, which applies penalties when significant prediction discrepancies occur among closely related students. By explicitly penalizing large prediction differences within neighborhoods defined by educational or task-specific similarity metrics (\textit{e.g.}, academic records, attendance patterns, learning styles), metric-consistency methods encourage stable and coherent model predictions. This helps reduce arbitrary distinctions at the individual level, thereby providing students equitable access to personalized interventions or resources. The theoretical foundations of metric-consistency originate from Dwork \textit{et al.}~\cite{Dwork2012Fairness}, who first conceptualized individual fairness through a Lipschitz condition emphasizing that similar individuals should receive similar outcomes, fairness in educational AI can be understood as ensuring consistency in learning opportunities and feedback among students with comparable abilities and efforts. Building on this foundation, Yurochkin \textit{et al.}~\cite{yurochkin2019training} extended this notion by introducing data-driven methods for learning fairness-aware similarity metrics, allowing educational AI systems, such as adaptive learning or recommendation platforms, to better capture relationships among students with comparable learning behaviors and performance levels, thereby ensuring equitable feedback and personalization across diverse learners. Similarly, Mukherjee \textit{et al.}~\cite{mukherjee2020two} developed practical approaches for integrating neighborhood-based consistency penalties explicitly into training objectives, promoting smoother predictions and improving fairness for similar individuals. Within educational AI contexts, metric-consistency provides a robust framework to ensure equitable treatment among students. By explicitly encoding similarity metrics relevant to educational tasks into the training process, metric-consistency methods effectively prevent arbitrary disparities, ensuring consistent predictions for students who exhibit comparable educational backgrounds and features.

\textbf{Ranking-based:} Ranking-based methods provide an alternative individual fairness approach by explicitly preserving the predictive ranking among students who share similar educational characteristics. Unlike methods that strictly enforce output similarity, these ranking-based approaches emphasize the relative ordering of students, ensuring that those with comparable educational backgrounds or performance levels receive consistent predictive rankings. To achieve this fairness goal, ranking-preserving techniques typically integrate constraints or regularization terms directly into the learning process. Specifically, these approaches construct a ranked list based on pairwise student similarities derived from their educational features (input space). A corresponding list is generated based on the predicted educational outcomes (output space). The fairness of the model is then evaluated by measuring the consistency between these two rankings. Ideal fairness is achieved when both lists perfectly align, meaning that the relative similarity of students remains stable through the model’s predictive process. Practically, fairness is assessed using ranking metrics such as Normalized Discounted Cumulative Gain (NDCG@$k$)~\cite{jarvelin2002cumulated} and Expected Reciprocal Rank (ERR@$k$)~\cite{chapelle2009expected}, which are commonly adopted tools for ranking-based fairness evaluations.

However, traditional ranking-based methods do not explicitly address scenarios involving censored outcomes, which frequently occur in educational data, such as incomplete graduation records or uncertain dropout predictions. To bridge this critical gap, Zhang \textit{et al.}~\cite{zhang2023individual} introduced Discounted Cumulative Fairness (DCF), an innovative fairness metric explicitly designed to combine ranking-based fairness evaluation with the ability to handle censored data. DCF extends conventional ranking approaches by directly incorporating censored survival information into fairness assessments, ensuring stable predictive rankings even when complete educational outcome data are unavailable. This methodological advancement is particularly relevant in educational contexts where data censorship is common due to student transfers, temporary dropouts, or extended enrollment periods that obscure definitive academic outcomes. Formally, DCF quantifies individual fairness by measuring the discounted cumulative similarity between rankings derived from censored predictions and the original input-based student rankings. Higher DCF scores reflect greater consistency and fairness in predictive outcomes amidst uncertainty. Further operationalizing this ranking-based fairness framework under censoring conditions, Zhang \textit{et al.}~\cite{zhang2023censored} proposed the Individual Fair Survival (IFS) algorithm, embedding DCF directly as a fairness constraint within survival analysis models. The IFS approach systematically ensures fair treatment and consistent predictive performance for students, even with censored data. Complementing this, Gillen \textit{et al.}~\cite{gillen2018online} demonstrated how ranking-based fairness constraints can be dynamically integrated into real-time student assessment and feedback systems. Similarly, Lahoti \textit{et al.}~\cite{lahoti2019operationalizing} operationalized individual fairness through explicit pairwise ranking comparisons, underscoring the practical importance of preserving relative student orderings. Collectively, these advanced ranking methods, with DCF prominently addressing the challenges of censorship, significantly enhance individual fairness evaluation in educational AI systems.

\textbf{Counterfactual Fairness:} Counterfactual fairness provides a causally informed pathway to achieving individual fairness: it requires models to maintain consistency in students' decisions under hypothetical scenarios—that is, by altering only students' sensitive attributes (\textit{e.g.}, race or gender) while keeping all other task-relevant features constant. The foundational work of Kusner \textit{et al.}~\cite{kusner2017counterfactual} formalized this notion through structural causal models, establishing that fairness is achieved when predictions in the factual and counterfactual worlds agree. Extending this idea to educational settings, recent research has used causal inference to uncover and mitigate latent biases in student data. For example, Kim \textit{et al.}~\cite{Kim2025CounterfactualFairnessEdu} demonstrated that counterfactual analysis on educational datasets can reveal subtle inequities in outcome predictions, such as models unfairly associating gender or race with academic success probabilities, which are often overlooked by conventional fairness metrics. On the methodological front, Kher \textit{et al.}~\cite{kher2025learning} proposed neural causal modeling techniques capable of generating realistic counterfactual samples, enabling models to learn fairness-aware representations without sacrificing predictive performance. These developments underscore the value of counterfactual fairness for educational AI applications, particularly in high-stakes decisions such as student admissions, scholarship allocation, or early-warning systems for dropout prediction. By ensuring that predictions remain invariant to demographic attributes, counterfactual fairness enables AI models to promote equitable educational opportunities and to support transparent, ethically grounded decision-making processes across diverse student populations.


\textbf{Pairwise Constraints:} Pairwise Constraints enforce individual fairness in educational AI by explicitly ensuring consistent treatment of students who share similar educational attributes or learning trajectories. Unlike methods that depend solely on continuous similarity metrics, this approach leverages explicit pairwise student relationships or constructs fairness-oriented graphs based on meaningful educational criteria, such as similar academic performance, coursework profiles, or collaborative interactions. In practice, this method typically integrates structured regularization terms or graph-based constraints, such as pairwise loss functions or graph Laplacian smoothing, into model training. These constraints directly penalize significant prediction disparities among students who are closely related, preventing arbitrary distinctions and ensuring fairer predictive outcomes at the individual level. In educational scenarios, pairwise and graph-based fairness constraints are particularly beneficial for applications involving collaborative or group learning contexts. For example, peer grading systems, collaborative learning platforms, and student performance prediction models can leverage fairness graphs to explicitly embed individual fairness considerations into their predictions. Lahoti \textit{et al.}~\cite{lahoti2019operationalizing} introduced a method that constructs pairwise fairness graphs derived directly from educational data, allowing predictive models to consistently treat students who share similar educational backgrounds and performance levels. Additionally, Kang \textit{et al.}~\cite{kang2020inform} presented a fairness-aware graph learning framework specifically suited for analyzing student networks, demonstrating its effectiveness in maintaining equitable treatment of interconnected learners, such as those collaborating on group projects or participating in peer assessment tasks. Collectively, these pairwise and graph-structured approaches provide robust methodologies to operationalize individual fairness in educational AI, ensuring that students with similar educational profiles are consistently and fairly treated by AI-driven educational interventions.

\subsection{Multi-level Fairness Methods}


Unlike the above works, which treat individual fairness and group fairness as distinct tasks, failing to consider the potential implications between them~\cite{Dwork2012Fairness,keya2021equitable,zhang2025fairness}, some recent methods integrate these perspectives. This separation of fairness objectives can unintentionally introduce additional biases, especially in scenarios such as education, where decisions deeply affect individual lives. Translating the running example in Figure~\ref{fig:figure1} to an educational setting, let $d_2$ be a student whose admission has already been confirmed, while the admission decisions for a male applicant $d_1$ and a female applicant $d_3$ are determined using a model constrained solely by individual fairness. In the input space (Figure~\ref{fig:figure1} (a)), these applicants share similar academic profiles and contextual features (such as GPA, coursework, and school background), evidenced by equal similarity distances: $(d_1,d_2)=(d_1,d_3)=(d_2,d_3)=3$. However, after applying individual fairness constraints alone (Figure~\ref{fig:figure1} (b)), the distances significantly change: $(d_1,d_2)=20$, $(d_1,d_3)=70$, and $(d_2,d_3)=90$. This shift occurs because the model enforces individual fairness constraints without considering group membership, leading to inconsistent constraints across groups—for example, a higher average constraint scalar for males compared to females. Consequently, the male applicant $d_1$, under a stronger individual fairness constraint, remains closer to $d_2$, resulting in admission, while the female applicant $d_3$, constrained more weakly, is distanced and consequently rejected. With the addition of group fairness awareness, the female applicant would instead be positioned at $d'_3$, closer to $d_2$, restoring equity by ensuring similar students receive similar educational opportunities, thus highlighting the importance of simultaneously addressing individual and group fairness to prevent unintended discrimination.

\begin{figure}[htpb] 
  \centering
  \includegraphics[width=0.95\textwidth]{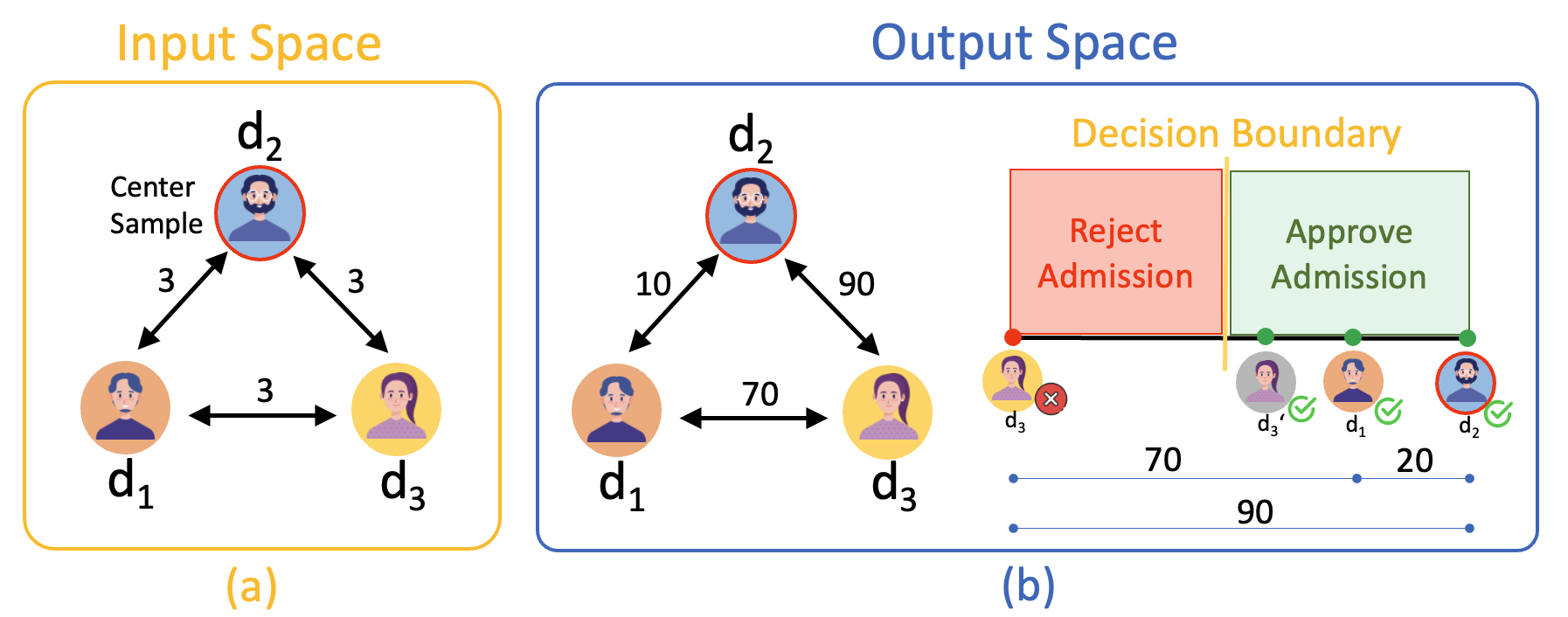} 
  \captionsetup{justification=centering} 
  \vspace{0.2cm}
  \caption{Example of unintended bias in educational AI when applying individual fairness constraints without considering group fairness, resulting in inequitable student admissions outcomes.} 
  \label{fig:figure1} 
\end{figure}

Motivated by these critical issues, Group- and Individual-aware Fair Censorship (GFIC)~\cite{wang2024group} integrates both group and individual fairness within a unified framework, introducing randomized distributions to ensure equitable ranking among individuals under group constraints. GIFC incrementally identifies an optimal probability distribution of rankings, ensuring that no individual disproportionately suffers from enforcing group fairness constraints. Applied to education, GIFC could equitably distribute resources such as scholarships or admission slots among students, ensuring fairness not only across demographic groups, but also at an individual level within each group. This helps educational institutions prevent scenarios in which a few students bear a disproportionate burden of group-level fairness adjustments. In a similar vein, Group Equality Individual Fairness (GEIF)~\cite{wang2024individual1} incorporates fairness constraints inspired by Rawlsian principles, enforcing a uniform distribution of individual fairness loss across demographic groups. By doing so, GEIF ensures that fairness adjustments do not systematically disadvantage specific students when educational opportunities are allocated. For example, in admissions or course placement decisions, GEIF ensures that each student experiences a comparable impact from fairness policies, effectively balancing fairness and merit considerations. Fair Graph U-Net~\cite{wang2025fair} further extends this integration of fairness into graph-based models, explicitly embedding fairness constraints into graph pooling mechanisms. Recognizing graph-based representations widely employed in educational scenarios, such as analyzing student collaborative networks or assessing peer-influence dynamics, Fair Graph U-Net ensures equitable student representation across demographic groups within the pooled representations. Moreover, it uses a probabilistic ranking approach to evenly distribute the fairness adjustments at the individual level, preventing particular students from becoming isolated or disadvantaged due to enforced group fairness constraints. This equitable treatment enhances fairness in downstream educational decisions, such as group assignments or tailored educational support. Lastly, Faircox~\cite{wang2024individual} combines group and individual fairness explicitly in ranking tasks by adjusting fairness constraints dynamically during the ranking optimization process. In educational settings, faircox can be directly applied to fairly rank students for awards, scholarships, or admissions based on their predicted academic success or potential contributions, ensuring fairness for both individual students and demographic groups simultaneously. By dynamically calibrating fairness constraints, faircox ensures that no student consistently bears a disproportionate loss from adjustments made to achieve group fairness, providing balanced educational opportunities aligned with both fairness and academic merit.

\section{Analytical Tools and Datasets}
\label{sec:tools_datasets}

Building upon the bias mitigation approaches discussed in the previous section, effectively applying these strategies in educational AI requires specialized analytical resources that connect theoretical frameworks to practical implementation. In this section, we introduce essential computational tools and publicly available datasets specifically designed to analysis and evaluate biases within educational contexts. The tools described provide methodological support for recognizing and mitigating biases in machine learning models commonly deployed in educational scenarios, while the datasets serve as empirical testbeds for validating fairness interventions and analyzing bias-related outcomes across diverse student populations.

\subsection{Analytical Tools}


Analytical tools play an indispensable role in educational AI fairness by enabling precise measurement, intuitive visualization, and detailed diagnosis of algorithmic bias. For instance, \textbf{AI Fairness 360 (AIF360)}~\cite{bellamy2018aif360} and \textbf{Fairlearn}~\cite{bird2020fairlearn} offer comprehensive capabilities for both evaluating fairness and deploying mitigation strategies. Specifically, AIF360 equips researchers and educators with extensive metrics and fairness-aware algorithms suitable for robust assessment of predictive fairness in educational scenarios, such as student outcome forecasting or resource allocation. Fairlearn complements this by providing intuitive visualizations and seamless Python integration, empowering practitioners to intuitively grasp fairness issues and readily implement mitigation methods in educational predictive models.

For interactive and scalable fairness evaluations, tools such as \textbf{TensorFlow Fairness Indicators}~\cite{tensorflow2020fairnessindicators} and Google's \textbf{What-If Tool (WIT)}~\cite{wexler2019whatif} stand out. TensorFlow Fairness Indicators is optimized for evaluating biases across large-scale predictive systems, particularly beneficial in educational contexts involving extensive student data or complex assessments. WIT, on the other hand, provides a highly interactive environment, allowing educators and researchers to visually explore model behavior and intuitively assess how predictions differ across various student demographics, making it ideal for examining subtle biases in grading or course recommendation systems.

In addition, \textbf{FairTest}~\cite{tramer2017fairtest} and \textbf{FairML}~\cite{Adebayo2016FairMLT} focus specifically on identifying and explaining the root causes of bias within predictive models. FairTest specializes in uncovering unintended correlations between sensitive student attributes and model outcomes, supporting researchers in pinpointing problematic associations in dropout predictions or automated grading. Complementing this, FairML provides feature-level interpretability, clarifying how demographic or contextual variables directly influence model predictions, thus improving transparency and guiding more equitable educational practices.

Furthermore, \textbf{Aequitas}~\cite{saleiro2019aequitasbiasfairnessaudit} specifically addresses accountability and transparency by rigorously auditing biases within binary classification scenarios common in education, such as admissions decisions or retention analyses. Through systematic identification and quantification of discrimination, Aequitas guides educational institutions in ensuring accountable, unbiased, and equitable AI-driven decisions. Collectively, these analytical tools equip educational practitioners and researchers with essential methodologies to rigorously identify, quantify, visualize, and ultimately mitigate biases, supporting the responsible and equitable deployment of AI in educational settings.

\begin{table}
\caption{Datasets Used in Fairness and Bias Research in AIED.}
\label{tab:datasets-detailed}
\centering
\renewcommand{\arraystretch}{1.1}            
\renewcommand\tabularxcolumn[1]{m{#1}}  
\footnotesize
\begin{tabularx}{\textwidth}{|
    >{\centering\arraybackslash}m{2.4cm}  
    |>{\centering\arraybackslash}m{1.6cm}  
    |>{\centering\arraybackslash}m{1.5cm}  
    |>{\centering\arraybackslash}m{1.5cm}  
    |>{\raggedright\arraybackslash}X       
    |>{\raggedright\arraybackslash}m{1.6cm}|} 
\hline
\textbf{Dataset} & \textbf{Categories} & \textbf{\#Samples} & \textbf{\#Features} & \textbf{Fairness Goal} & \textbf{Sensitive Attributes} \\
\hline
\hline
TIMSS~\cite{Mullis2016TIMSS} & IALS & $>$300,000 students & $>$200 & Fairness across educational systems & Country, gender, educational resources \\
\hline
PISA~\cite{OECD2019PISA} & IALS & $>$500,000 students & $>$600 & Fairness across socioeconomic contexts & Country, gender, socioeconomic status \\
\hline
NELS:88~\cite{Ingels2005NELS} & IALS & $>$25,000 students & $>$1,000 & Educational predictions fairness & Race, gender \\
\hline
ELS:2002~\cite{bozick2007education} & IALS & 15,000 students & $>$1,000 & Educational trajectories improvement & Race, gender, school type \\
\hline
PERSUADE 2.0~\cite{crossley2024large} & IALS & $>$25,000 essays & 14 & Scalable scoring fairness analyses & Race, gender \\
\hline
EdNet~\cite{Choi2020EdNet} & ILAD & $>$100M interactions & Multiple & Personalized learning fairness & Age, gender, locations \\
\hline
KDDCUP 2015~\cite{feng2019understanding} & ILAD & 180,785 study logs & Multiple & Dropout prediction fairness & Gender, locations \\
\hline
OULA~\cite{Kuzilek2017OULA} & ILAD & $>$10M interactions & Multiple & Learning scores prediction fairness & Gender, disability status \\
\hline
TutOR~\cite{Stamper2006TutOR} & ILAD & Varies & Multiple & Adaptive learning system development & Age, gender \\
\hline
MOOCdb~\cite{Veeramachaneni2014MOOCdb} & ILAD & Varies & Multiple & Engagement and learning behavior analysis & Location, educational background \\
\hline
CMU DataShop~\cite{Koedinger2012CMUDataShop} & ILAD & Varies & Multiple & Instructional strategy evaluation & Age, gender, race, disability status  \\
\hline
LAK~\cite{Siemens2012LAK} & ILAD & Varies & Multiple & Academic performance and behaviors analysis & Educational level, discipline \\
\hline
CBT~\cite{Hill2016CBT} & ILAD & Varies & Multiple & Literacy development study & Age, culture background \\
\hline
ASSISTments~\cite{Feng2009ASSISTments} & ILAD & $>$100,000 interactions & Multiple & Socioeconomic and gender impact study & Gender, socioeconomic status  \\
\hline
CCD~\cite{glander2017documentation} & PIFD & $>$100,000 schools & $>$500 & Equitable education policy support & Race, English proficiency, disability status \\
\hline
IPEDS~\cite{broyles1994integrated} & PIFD & $>$7,000 institutions & $>$200 & Transparent and equitable post-secondary access & Race, gender, financial aid status \\
\hline
Aequitas WP7~\cite{giovanelli_2025_15546688} & PIFD & 80,174 students & 153 & Equal educational access and outcomes & Gender, location \\
\hline
\end{tabularx}
\end{table}

\subsection{Datasets}

Datasets provide essential resources for fairness analysis in educational AI research by offering standardized contexts to evaluate, detect, and mitigate biases in diverse educational settings. The effectiveness of fairness assessments and bias mitigation techniques is strongly influenced by dataset features such as demographic coverage, granularity, and the scope of captured educational interactions. As summarized comprehensively in Table~\ref{tab:datasets-detailed}, fairness-oriented educational research primarily utilizes three broad categories of benchmark datasets: \textbf{(1) International Assessments and Longitudinal Studies (IALS)}, \textbf{(2) Interaction-based Learning Analytics Datasets (ILAD)}, and \textbf{(3) Policy and Institutional Fairness Datasets (PIFD)}. Each category provides unique strengths and analytical contexts that significantly shape fairness research methods and outcomes.

\textbf{International Assessments and Longitudinal Studies (IALS)}, such as TIMSS, PISA, NELS:88, ELS:2002 and the newly included PERSUADE corpus, provide large-scale standardized measures of educational performance across diverse cultural, socioeconomic, and demographic contexts. These datasets typically incorporate extensive student, teacher, and institutional background information, allowing researchers to conduct cross-national or longitudinal fairness analyzes. For example, TIMSS and PISA facilitate comparative studies of educational equity internationally, allowing the exploration of systemic biases across educational policies and socioeconomic conditions. Similarly, longitudinal datasets such as NELS:88 and ELS:2002 uniquely support analyses of fairness across extended academic trajectories, capturing cumulative effects of demographic disparities on educational outcomes over time. Additionally, specialized datasets such as the PERSUADE corpus facilitate fine-grained analyses of measurement bias in automated essay scoring, examining disparities in writing evaluation across demographic and socioeconomic subgroups.

\textbf{Interaction-based Learning Analytics Datasets (ILAD)}, including EdNet, KDD Cup 2015, OULAD, TutOR, MOOCdb, and CMU DataShop, capture detailed student interactions within online educational environments and Intelligent Tutoring Systems (ITS). Unlike international assessment datasets, these interaction-based datasets allow for high-resolution analysis of biases at the individual learner and adaptive-system levels. EdNet and KDD Cup 2015, for instance, contain detailed logs of student interactions, enabling fairness-focused research into personalization biases, dropout prediction accuracy, and adaptive learning effectiveness across demographic subgroups. Furthermore, datasets such as OULAD and MOOCdb facilitate fine-grained analysis of how interaction patterns differ among students from diverse demographic backgrounds, helping to identify subtle biases embedded in AI-driven personalized education systems.

\textbf{Policy and Institutional Fairness Datasets (PIFD)} provide institutional level and structural demographic data such as CCD and IPEDS, suitable for examining fairness concerns from the allocation of policy-oriented and resource perspectives. Distinctively different from individual-centric interaction datasets, these datasets focus on institutional demographics, enabling macro-level analyses of educational equity, policy effectiveness, and institutional resource distribution fairness. For example, CCD encompasses comprehensive annual demographic information for all U.S. public schools, providing foundational data for policy researchers examining disparities in school resources, funding, and educational access. Similarly, the Aequitas WP7 dataset, explicitly designed as a fairness-oriented benchmark, further enhances this category by providing structured frameworks for systematically evaluating the fairness impacts of educational policies and institutional interventions in various demographic contexts. Similarly, IPEDS contains detailed institutional-level information on enrollment, financial aid, and completion rates within postsecondary education, offering insights into equity issues related to access to higher education and institutional performance.

\section{Ethical Considerations and Policy Guidelines for AI in Education}
\label{sec:ethical_considerations}

The challenges facing AI in education are not only technical in nature, but also deeply ethical. This section examines key ethical dilemmas specifically associated with ensuring fairness, equity, and privacy within AI-driven educational systems. It underscores the importance for researchers, practitioners, and policymakers to possess a nuanced understanding of ethical principles guiding fair and equitable AI applications. In addition, we provide a comprehensive overview of global ethical guidelines, standards, and policy frameworks designed to guide the responsible integration of AI into education, thereby promoting inclusive and equitable educational outcomes.

\begin{table}
\caption{Guidelines, Policies and Regulations Addressing Fairness in AIED.}
\label{table:fairness-AIED}
\renewcommand{\arraystretch}{1.7}
\begin{tabular}{|p{1.5cm}|p{3.8cm}|p{3.0cm}|p{1.8cm}|p{0.7cm}|}
\hline
\textbf{Type} & \centering\textbf{Document } & \textbf{Organization} & \textbf{Country} & \textbf{Year} \\
\hline
\hline
Guidelines & Ethics Guidelines for Trustworthy AI~\cite{cannarsa2021ethics} & European Commission & European Union & 2021 \\
\hline
Guidelines & UNESCO's Recommendations on the Ethics of AI~\cite{unesco2022recommendation} & United Nations Educational, Scientific and Cultural Organization (UNESCO) & International & 2022 \\
\hline
Guidelines & Ethical Guidelines for Educators on Using AI and Data in Teaching and Learning~\cite{ec_educators_ai_2022} & European Commission & European Union & 2022 \\
\hline
Guidelines& A Toolkit for Safe, Ethical, and Equitable AI Integration~\cite{edtoolkit_nondiscrimination_2023} & U.S. Department of Education & United States & 2023 \\
\hline
Guidelines & Ethical Guidelines for Artificial Intelligence in Higher Education~\cite{educause_ai_guidelines_2025} & Educause & United States & 2025 \\
\hline
Policies & Montreal Declaration for a Responsible Development of Artificial Intelligence~\cite{montreal_declaration_2018} & Université de Montréal & Canada & 2018 \\
\hline
Policies & OECD Principles on Artificial Intelligence~\cite{oecd_ai_principles_2019} & Organization for Economic Co-operation and Development (OECD) & International & 2019 \\
\hline
Policies & AI in Education: Guidance for Policy-Makers~\cite{unesco_ai_policy_2021} & United Nations Educational, Scientific and Cultural Organization (UNESCO) & International & 2021 \\
\hline

Policies & National AI Strategy~\cite{uk_national_ai_strategy_2021} & UK Government & United Kingdom & 2021 \\
\hline

Policies & Guidance for Developing Policies to Govern the Adoption and Use of Artificial Intelligence in K-12 Schools~\cite{ross2024guidance} & U.S. Governments & United States & 2024 \\
\hline
Policies  & Task Force Report and Proposed Policy Statement on AI in Education~\cite{nea_ai_taskforce_2024} & National Education Association & United States & 2024 \\
\hline
Policies & Advancing Artificial Intelligence Education for American Youth~\cite{whitehouse_eo_ai_2025} & U.S. The White House & United States & 2025 \\
\hline
Regulations & AI for Good~\cite{itu_ai_for_good_2017} &  International Telecommunication Union (ITU) & International & 2017 \\
\hline
Regulations & Dear Colleague Letter on AI Use in Schools~\cite{ed_dept_dcl_ai_2025} & U.S. Department of Education & United States & 2025 \\
\hline
\end{tabular}
\end{table}

\subsection{Ethical Dilemmas}

Ethical education in AI aims to foster a profound understanding of the societal impacts of AI technologies in educational contexts. A central ethical challenge lies in addressing bias and fairness within AI-driven educational algorithms. As highlighted in \cite{ONeil2016Weapons}, without careful design and rigorous review, algorithms may inadvertently perpetuate or amplify existing social biases, leading to discriminatory outcomes that disproportionately affect historically marginalized student groups. This underscores the necessity of integrating critical thinking about ethical principles into the design, deployment, and evaluation of AI systems specifically within educational settings. In addition, privacy and data protection constitute critical ethical concerns in educational AI. AI-based educational systems often require large datasets containing sensitive student information, raising critical ethical questions about data ethics, informed consent, and privacy compliance~\cite{TaylorFloridiSloot2017}. Educational AI practices require rigorously adhere to privacy laws and ethical guidelines to protect students' rights and ensure that the benefits of AI-driven personalization and intervention do not compromise student privacy or reinforce existing inequalities. Therefore, effective addressing these ethical dilemmas is essential to promoting equitable educational outcomes and ensuring that AI integration genuinely benefits all students, particularly those from disadvantaged backgrounds.

\subsection{Guidelines, Policies, and Regulations}
The landscape of guidelines, policies, and regulations for AIED, summarized in Table~\ref{table:fairness-AIED}, reflects a growing global commitment to ethical, equitable, and inclusive AI practices. Ethical guidelines, such as the United Nations Educational, Scientific and Cultural Organization (UNESCO) Recommendations on the Ethics of AI and the European Commission Ethics Guidelines for Trustworthy AI, underscore transparency, accountability, fairness, and inclusiveness as critical factors in the ethical deployment of AI technologies within educational environments. The U.S. Department of Education's A Toolkit for Safe, Ethical, and Equitable AI Integration similarly provides detailed ethical considerations to assist stakeholders in navigating fairness, equity, and privacy issues specific to educational contexts. Furthermore, the European Commission's Ethical Guidelines for Educators explicitly address the ethical use of AI and data within teaching and learning settings.

Complementing these ethical frameworks are policy initiatives aimed at shaping practical implementation. UNESCO's AI in Education guidance for policymakers, the Organization for Economic Cooperation and Development (OECD) Principles on Artificial Intelligence, and the Montreal Declaration by Université de Montréal all emphasize inclusive development, fairness, and the responsible advancement of AI. National-level policies, such as the UK's National AI Strategy and the U.S. government's recent initiative, Advancing Artificial Intelligence Education for American Youth, highlight the necessity of cultivating AI literacy and equitable educational opportunities. Meanwhile, state-level guidelines in the U.S. and institutional policies, such as the NEA’s Task Force Report and Policy Statement, further reinforce ethical and equitable AI adoption at local and organizational levels. To ensure compliance and practical implementation, formalized regulations have also emerged, including the International Telecommunication Union (ITU)'s AI for Good initiative and the U.S. Department of Education’s Dear Colleague Letter on AI Use in Schools. Collectively, these updated documents establish a comprehensive framework guiding stakeholders toward the ethical, equitable, and transparent use of AI in education, thus supporting inclusive educational practices and equitable outcomes.


\section{Challenges and Future Directions }
\label{sec:challenges_fd}
Despite substantial progress in understanding algorithmic fairness, mitigating biases, and addressing ethical considerations in educational AI, numerous challenges and open questions remain regarding the equitable integration of AI in educational contexts. In this section, we identify key unresolved issues and outline promising avenues for future research, aiming to inspire further interdisciplinary efforts toward fairer and ethically responsible educational AI systems.

\textbf{Personalization versus data privacy:} Given the widespread use of personalized learning, more research is needed on how AI may be most effectively utilized to promote personalized learning. This includes tasks such as proposing learning materials tailored to individual learners and providing guidance in problem solving~\cite{chen2021twenty}. However, in order to provide customized experiences, AI model training necessitates the use of large-scale learner data, which may include extremely sensitive information. Occasionally, models unintentionally retain training data that contain sensitive information, which becomes exposed during the analysis of the model. However, the full capabilities of a machine learning model can only be realized by examination of student data~\cite{chan2019applications}. Due to the inability of most existing models to ensure that their output can be applied to different learners beyond specific individual characteristics, as well as the potential risk to learners' data and the negative impact on the acceptance of AI in society, it is necessary to restrict instructors' access to learners' data in order to appropriately limit learners' exposure to instructors' knowledge. Educational institutions should practice transparency on student data privacy to address any misconceptions and concerns related to data use.

\textbf{Digital divide in AI integration education:} The digital divide represents a substantial challenge to equitable AI integration within education. Previous studies have consistently emphasized how disparities in access to digital infrastructure, devices, and reliable internet connectivity can further aggravate existing inequalities, particularly in developing countries~\cite{Asher2021COVID19DL}. For example, recent research by Veras \textit{et al.}\cite{veras2024artificial} highlights specific barriers within physiotherapy education, underscoring the urgency of addressing such technological gaps. To effectively mitigate the impact of digital inequalities on student outcomes, future work requires prioritize ensuring equitable access to critical technologies, internet resources, and targeted training programs\cite{Iftikhar2023ANALYZINGDD}. Additionally, ethical considerations are fundamental to the responsible development and deployment of educational AI solutions. Achieving sustainable progress in closing the digital divide requires coordinated collaboration among governmental bodies, educational institutions, industry stakeholders, and civil society organizations~\cite{Suleiman2023BridgingTD}.

\textbf{Teacher and student adoption:} The adoption of AI technologies by teachers and students in education is a complex challenge requiring targeted strategies for effective implementation. Research emphasizes the importance of developing adaptive learning systems tailored to individual learning styles, necessitating comprehensive teacher training for the successful integration of AI into pedagogical practices~\cite{Ragab2010AACWELSA}. Although interactive language tools and virtual platforms provide personalized learning experiences, teachers need to skillfully guide students to maximize the benefits of these technologies~\cite{Alam2021ShouldRR}. Moreover, students' metacognitive awareness significantly influences their engagement with AI tools such as ChatGPT, underscoring the need for customized support and training~\cite{Abdelhalim2024UsingCT}. Future initiatives should establish extensive training programs for educators and students, including workshops, online resources, and collaborative learning communities, to facilitate sharing best practices and addressing emerging challenges~\cite{Khalil2023RoleOA}. 

\textbf{Global Collaboration and Standardization:} A critical future direction is to foster international collaboration aimed at developing standardized guidelines and best practices for ensuring fairness in educational AI. By establishing global partnerships, educational institutions and researchers can effectively harmonize their fairness efforts, share essential resources such as datasets, analytical tools, and evaluation methods, and leverage diverse global expertise to systematically address common fairness challenges~\cite{Vinuesa2020}. This approach can accelerate progress towards equitable AI-driven education, promoting consistency and transparency across international educational contexts.

\nocite{wang2023preventing,zhang2023individual,wang2023fg2an,wang2023mitigating,chinta2023optimization,chu2024fairness,yin2024improving,chinta2024fairaied,wang2024individual1,doan2024fairness1,wang2024advancing,wang2024group,wang2024individual,yin2024accessible,wang2025fg,wang2025graph,wang2025fair,wang2025towards2,yin2025digital,chinta2025ai,wang2025fdgen,wang2025Fairness,wang2025Redefining,zhang2019faht,zhang2024ai,zhang2022longitudinal,zhang2023censored,zhang2025fairness,zhang2022fairness,wang2024towards1,saxena2023missed,zhang2019fairness,zhang2020flexible,zhang2020online,zhang2020learning,zhang2021farf,zhang2021fair,zhang2023fairness,zhang2016using,zhang2018content,zhang2021autoencoder,zhang2018deterministic,tang2021interpretable,zhang2021disentangled,yazdani2024comprehensive,liu2021research,liu2023segdroid,cai2023exploring,guyet2022incremental,zhang2024fairness,wang2025FairnessT,zhang2025online,yinAMCR2025,Wang2025Unified,ijcai2025p64,ijcai2025p63,zhang2025datasets,palikhe2025towards,yin2025Uncertain,amon2024uncertain,wang2025ai}

\section{Conclusion}
\label{sec:conclusion}

In this survey, we presented a comprehensive review of fairness, bias, and ethical considerations in educational AI by systematically following our proposed taxonomy categorizing biases into individual-level, group-level, and multi-level dimensions. Guided by this unified analytical framework, we systemically analyzed fairness definitions, bias sources, and mitigation strategies in educational contexts. Our survey highlights several critical yet overlooked challenges, including balancing personalization and data privacy, bridging the digital divide, and facilitating effective adoption by educators and students. By emphasizing interdisciplinary collaboration, this survey outlines future directions aimed at developing comprehensive fairness frameworks and fostering cross-sectoral cooperation. Ultimately, we advocate for sustained joint efforts among researchers, educators, and policymakers to ensure ethically responsible and equitable AI integration in education.
 
\bibliography{sn-bibliography}

\end{document}